\PassOptionsToPackage{dvipsnames,dvipsnames,svgnames,x11names}{xcolor}
\documentclass[11pt,a4paper]{article}
\usepackage{times,latexsym}
\usepackage{url}
\usepackage[T1]{fontenc}

\usepackage[acceptedWithA]{tacl2021v1}
\usepackage{graphicx}
\usepackage[most]{tcolorbox}
\usepackage{array}
\usepackage{booktabs}
\usepackage{tabularx}
\usepackage{placeins}
\usepackage[dvipsnames,svgnames,x11names]{xcolor}
\usepackage{colortbl}
\usepackage{multirow,makecell}
\usepackage{cleveref}
\usepackage{epigraph}
\usepackage{pifont}
\usepackage{ragged2e}
\usepackage{arydshln}
\usepackage[most]{tcolorbox}
\usepackage{tikz}
\usetikzlibrary{decorations.shapes}
\usepackage{amsmath}
\usepackage{amssymb}
\usepackage[normalem]{ulem}
\usepackage{hypcap}
\usepackage{capt-of}
\usepackage{fontawesome}
\usepackage{bm}
\usepackage{romannum}
\usepackage{atbegshi}
\usepackage{hyperref}
\usepackage{colortbl}
\usepackage{subfigure}

\definecolor{OliveGreen}{rgb}{0.0, 0.5, 0.0}
\definecolor{ggreen}{rgb}{0.0, 0.6, 0.0}
\definecolor{rred}{rgb}{0.75, 0.0, 0.0}
\definecolor{bblue}{rgb}{0.13, 0.67, 0.8}
\definecolor{sblue}{RGB}{240,248,255} 
\definecolor{sred}{RGB}{255,240,245} 
\definecolor{sgreen}{RGB}{239, 255, 232} 

\newcolumntype{C}{>{\arraybackslash}X}
\newcommand{\greencheck}{\textcolor{LimeGreen}{\ding{52}}}
\newcommand{\redcross}{\textcolor{Red2}{\ding{56}}}
\newcommand{\ourbenchmark}{\textsc{LexBench}}
\newcommand{\hlcell}{\cellcolor{AliceBlue!80}}
\newcommand{\hlcelll}{\cellcolor{LavenderBlush!80}}
\newcommand{\hlcellll}{\cellcolor{LightGreen!15}}
\NewDocumentCommand{\fullcite}{ m }{%
    \citeauthor{#1}, \citeyear{#1}%
}

\hypersetup{colorlinks=true, urlcolor=OliveGreen}

\tikzset{
    decoration = {shape backgrounds, shape=circle, shape size=0.8pt, shape sep={2pt, between centers}},
    paint/.style = {decorate, fill=black}
}
\newtcolorbox{DottedTextBox}{
    enhanced,
    boxsep=0mm, left=4pt,
    arc=4pt,
    colback=white,
    colframe=white,
    pad at break=0pt,bottomrule at break=0pt,toprule at break=0pt,
    borderline={0pt}{0pt}{paint},
}

\usepackage{xspace,mfirstuc,tabulary}

\newif\iftaclinstructions
\taclinstructionsfalse 
\iftaclinstructions
\renewcommand{\confidential}{}
\renewcommand{\anonsubtext}{(No author info supplied here, for consistency with
TACL-submission anonymization requirements)}
\newcommand{\instr}
\fi

\iftaclpubformat 

\else

\fi

\title{Revisiting\textit{ a Pain in the Neck}:\\Semantic Phrase Processing Benchmark for Language Models}

\author{Yang Liu$^{\dag}$, ~~Melissa Xiaohui Qin$^{\ddag}$, ~~Hongming Li$^{\dag}$, ~~Chao Huang$^{\dag}$\thanks{~~Corresponding author} \\
   $^{\dag}$School of Computer and Communication Engineering, $^{\ddag}$School of Foreign Studies \\
    University of Science and Technology Beijing, Beijing, China \\
  \texttt{\{yangliu.real,dearmelissaqin\}@gmail.com} \\
  \texttt{chaohuang@ustb.edu.cn} 
 }

\date{}

\begin{document}
\maketitle
\begin{abstract}
We introduce \textbf{\ourbenchmark}, a comprehensive evaluation suite enabled to test language models (LMs) on ten semantic phrase processing tasks. Unlike prior studies, it is the first work to propose a framework from the comparative perspective to model the general semantic phrase (i.e., lexical collocation) and three fine-grained semantic phrases, including idiomatic expression, noun compound, and verbal construction. Thanks to \ourbenchmark, we assess the performance of 15 LMs across model architectures and parameter scales in classification, extraction, and interpretation tasks. Through the experiments, we first validate the scaling law and find that, as expected, large models excel better than the smaller ones in most tasks. Second, we investigate further through the scaling semantic relation categorization and find that few-shot LMs still lag behind vanilla fine-tuned models in the task. Third, through human evaluation, we find that the performance of strong models is comparable to the human level regarding semantic phrase processing. Our benchmarking findings can serve future research aiming to improve the generic capability of LMs on semantic phrase comprehension. Our source code and data are available at \faicon{github} \url{https://github.com/jacklanda/LexBench}.
\end{abstract}

\section{Introduction}\label{sec:introduction}
\epigraph{{\textit{``You shall know a word by the company it keeps.''}}}{----- \fullcite{philological1957studies}}\vspace{-0.5em}
Semantic phrases (SPs), also known as multiword expressions (\textsc{MwE}), are word combinations idiosyncratic concerning syntax or semantics \cite{pasquer-etal-2020-verbal}. They have been well explored in taxonomy and categorized into different types by their semantic relation of lexical combination, i.e., the lexical function (LF) \cite{mel1998collocations}. Furthermore, semantic phrases can be categorized by their varying compositionality, idiomaticity, polylexicality, and fixedness \cite{sailer2018multiword}. A fine-grained taxonomy of semantic phrases from a compositional perspective includes idiomatic expressions, noun compounds, and verbal constructions \cite{ramisch2023multiword}. The issue of semantic phrase handling is crucial for NLP systems, where it raises many challenges \cite{constant2017multiword}, making them ``a pain in the neck'' for NLP applications in a long time \cite{sag2002multiword,shwartz-dagan-2019-still}. Hence, relevant tasks of semantic phrase processing have been the focus of numerous research works \cite{ramisch2023survey,wada-etal-2023-unsupervised,tanner-hoffman-2023-mwe}. In this paper, we concentrate on the four representative phenomena of semantic phrases, which are categorized by their various degrees of compositionality in the majority: idiomatic expression, noun compound, verbal construction, and lexical collocation.

\begin{figure*}[t]
    \centering
    \includegraphics[width=0.8\linewidth]{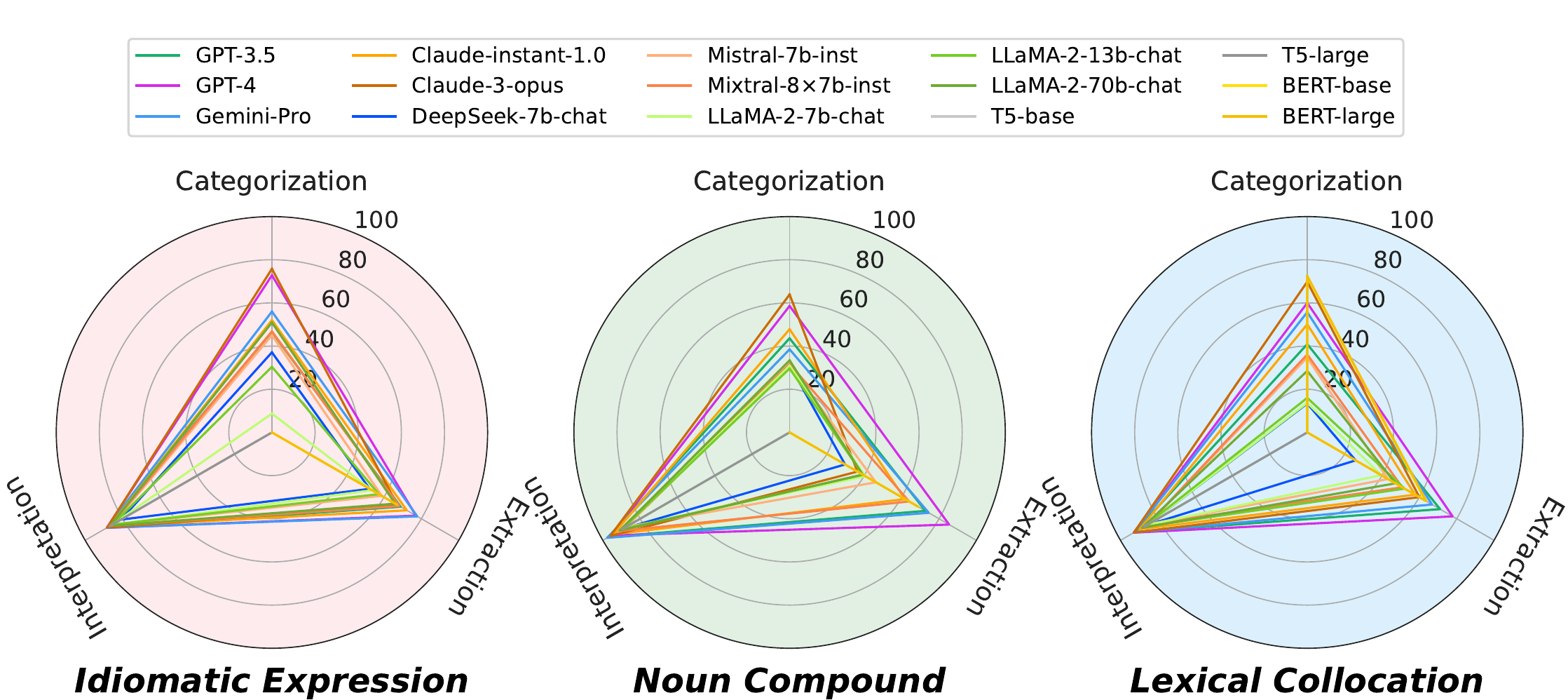}
    \caption{Overall the best performance (i.e., capacity triangle $\triangle$) of models on \ourbenchmark}
    \label{fig:Performance-SpiderFigure}
\end{figure*}

Idiomacity Expressions (henceforth: IE), as typical non-compositional (a.k.a. semantic idiomaticity) phrase, e.g., \textit{kick the bucket} and \textit{bite the bullet}, are of particular interest in that their meaning cannot be obtained by compositionally interpreting their word constituents. It raised increasing attention in the work of recent years \cite{zhou2022idiomatic,zeng-bhat-2022-getting,chakrabarty2022s,haviv-etal-2023-understanding}.

Verbal constructions (henceforth: VC), a.k.a., verbal multiword expressions (\textsc{VMwE}), have been the focus of the \textsc{MwE} community \cite{tanner-hoffman-2023-mwe,savary-etal-2023-parseme,ramisch-etal-2023-survey}. Most of VC are half-compositional (or partly compositional) phrase, the incomplete semantic of each VC can be concluded by its partial words. The common verb constructions comprise light-verb constructions (LVC), verb-particle construction (VPC), and verbal idioms (VID), i.a.

Noun compounds (henceforth: NC) are also frequent lexical combination phenomenon. The bulk of NC can be categorized as fully compositional phrases. Implicit meanings in NC often necessitate prior world knowledge for comprehension. For example, that \textit{sweet} refers to the \textit{flavor} of \textit{candy} but to the \textit{status} of \textit{victory} \cite{kolluru-etal-2022-covid,lyu-etal-2022-favorite,coil-shwartz-2023-chocolate}.

Lexical collocation (henceforth: LC) \cite{mel1995phrasemes} is a broad semantic phrase type. In this paper, we argued that lexical collocations constitute a superset of other semantic phrase types. Each instance of LC can be idiosyncratic to various degrees in compositionality. For example, the meaning of \textit{heavy rain} is similar to the sum of its components, while \textit{sit on (a) jury} is less compositional, even some collocations (e.g., \textit{get cold feet}) behave more idiomaticity. It is usually classified by fine-grained taxonomy in terms of lexical functions \cite{wanner1996lexical}, and is explored to perform many tasks recently \cite{espinosa-anke-etal-2016-extending,rodriguez-fernandez-etal-2016-semantics,espinosa-anke-etal-2019-collocation,espinosa-anke-etal-2021-evaluating,math10203831}.

Recently, large language models (LLM) have significantly attained advanced improvement in model capacity \cite{openai2023gpt4,geminiteam2023gemini,jiang2023mistral,touvron2023llama,TheC3}. To assess the performance of models, a standard method involves creating evaluation benchmarks to gauge the proficiency of LMs across various aspects \cite{chang2023survey,minaee2024large}, such as general language understanding \cite{hendrycks2020measuring,huang2023ceval}, mathematics \cite{Cobbe2021TrainingVT,patel-etal-2021-nlp}, reasoning \cite{Suzgun2022ChallengingBT,Srivastava2022BeyondTI} and code generation \cite{Chen2021EvaluatingLL,Austin2021ProgramSW}. Especially, GPT-4 is considered the ``Sparks of Artificial General Intelligence'' \cite{bubeck2023sparks} and exhibits remarkable capabilities across domains and tasks.

Although many tasks have been proven successful based on LMs, the challenging comprehension and processing tasks on semantic phrases remain an open question. Hence, we can naturally file the following research questions: ``(\romannum{1}) Can (large) language models perform well in semantic phrase processing tasks?'', and if they can, the second question is ``(\romannum{2}) To what extent can LMs be well-performing in these tasks? (Does semantic phrase processing still be a pain in the neck?)''. Meanwhile, as pointed out recently by \cite{miletic2024semantics}, there currently exists a need for directly comparable evaluation framework to encompass comprehensive assessment of different semantic phrase phenomena.

To answer these questions and bridge the gaps. We propose a suite of evaluation framework,  \ourbenchmark, which allows us to empirically compare GPT-4, Claude-3-Opus, and current strong LLMs with in-context learning in SP processing. In addition, we run local models to report several recent works such as Llama-2 \cite{touvron2023llama2}, DeepSeek-7B \cite{bi2024deepseek}, Mistral-7B \cite{jiang2023mistral}, and Mixtral-8x7B \cite{jiang2024mixtral}. Furthermore, we also train and test several small language models (SLM), such as BERT \cite{devlin2019bert} and T5 \cite{raffel2020exploring}, with supervised fine-tuning methods.

To our knowledge, \ourbenchmark\ is the first and the largest public benchmark designed for neural SP processing with LMs, aiming to provide a profile the performance in semantic phrase comprehension and processing with LMs. Overall, \textbf{our contributions} are summarized threefold:
\begin{enumerate}
  \item \textbf{Novel Benchmark.} We propose a comprehensive benchmark to evaluate a wide range of LMs on real-world semantic phrase processing applications and bridge a crucial research gap. Drawing on existing data sources, we compile ten new datasets and tailor a series of auto metrics to provide a solid testbed and toolkit for the \textsc{MwE} community.
  \item \textbf{Scaling Analysis.} We pioneer in investigating how model scales and in-context learning can boost the capability of LLMs to diverse phrase and task formats, especially in the effect of scaling semantic relation categorization. Meanwhile, via the analysis of prompt strategy, we propose \textsc{Oracle Prompting}, a recipe to elicit semantic phrase extraction with the guidance of the Oracle definition.
  \item \textbf{Empirical Findings.} Thanks to \ourbenchmark, we unveil the pros and cons of current LMs. Further performance enhancement in semantic phrase processing remains a challenging problem, especially in the extraction and classification task formats. Through human evaluation, we find that the capacity of LMs still needs to be improved, although beyond humans in many tasks. To some extent, it is still ``a pain in the neck'' in grounding even to the most potent NLP systems currently.
\end{enumerate}

\begin{figure}[t]
    \centering\hspace{-1em}
    \includegraphics[width=0.85\linewidth]{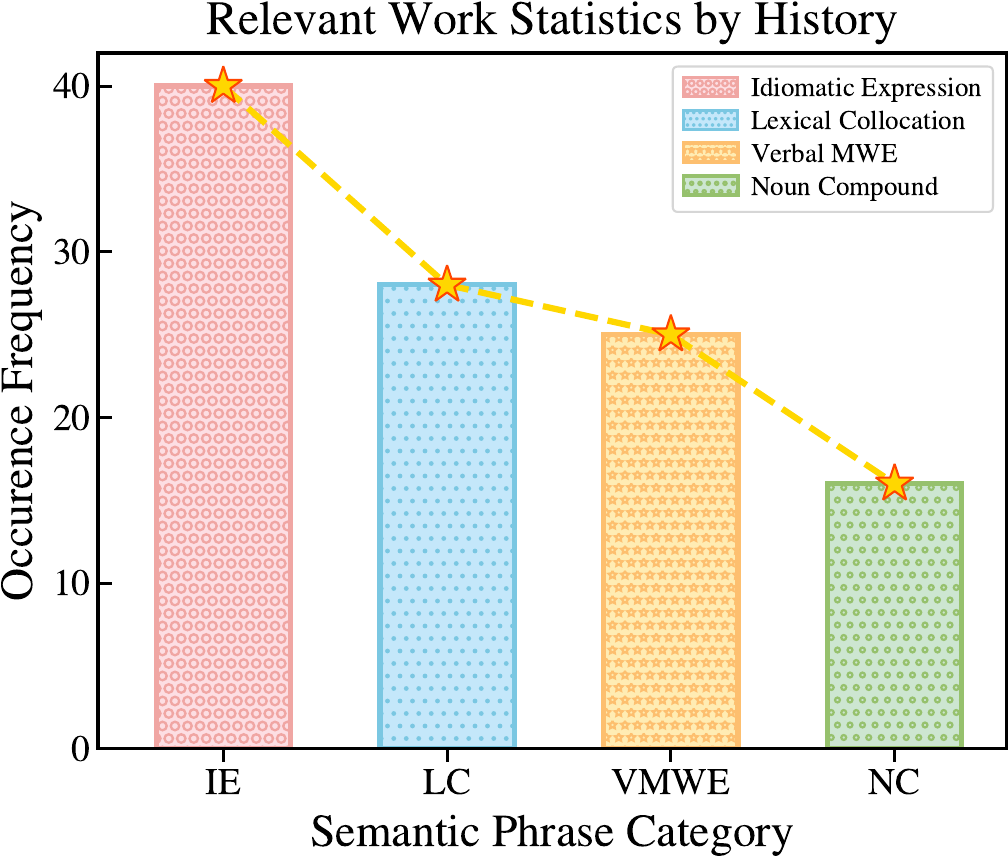}
    \caption{We manually curated the works related to semantic phrase processing in NLP published from 2010 to the present in Google Scholar and presented them in descending order (from left to right).}
    \label{fig:Relevant-work-statistics}
    \vspace{-0.8em}
\end{figure}

\section{Related Work}

\paragraph{Semantic Phrase Processing}
An extensive body of work exists on the evaluation of semantic phrase processing in context and also increasingly on different tasks \cite{vacareanu-etal-2020-unsupervised,arase-tsujii-2020-compositional,klubicka-etal-2023-idioms,wada-etal-2023-unsupervised}, even though not yet introducing the mainstream paradigm of LLMs \cite{pham-etal-2023-pic,buijtelaar-pezzelle-2023-psycholinguistic,zeng-etal-2023-iekg}. \cite{ponkiya-etal-2020-looking,shwartz-2021-long,coil-shwartz-2023-chocolate} shown that both few-shot LLMs and supervised fine-tuned T5 models can be well-performing in noun compound paraphrasing. However, larger transformer-based pre-trained models (e.g., GPT-4) with their in-context learning characteristics have not yet been used widely for tasks related to multiword expressions.

\paragraph{Semantic Phrase Resource}
For broad semantic phrases, the DiMSUM shared task \cite{schneider-etal-2016-semeval} is the first popularized to perform lexical semantic unit detection in \textit{sequence labeling} scheme. Lexical collocation, as addressed by the collection \textsc{LexFunc} \cite{espinosa-anke-etal-2019-collocation}, was presented to classify collocation via word embeddings. Subsequently, it was expanded to facilitate contextual collocate retrieval and collocation categorization \cite{espinosa-anke-etal-2021-evaluating}. Concurrently, multilingual collocation resources were proposed, such as \cite{fisas-etal-2020-collfren,espinosa-anke-etal-2022-multilingual,nisho2022extraction}. Many datasets have also been issued for idiomatic expression to support tasks \cite{Tedeschi2022ID10MII,zhou-etal-2021-pie,haagsma-etal-2020-magpie}. Among them, \cite{Tedeschi2022ID10MII} presents a multilingual corpus including ten languages, while \cite{zheng-etal-2019-chid} presents a large dataset for cloze test on idioms in Chinese. For verbal construction, \cite{kato-etal-2018-construction,ramisch-etal-2018-edition,ramisch-etal-2020-edition,savary-etal-2023-parseme} focus on verbal \textsc{MwE}, such as \textit{to take a decision}, \textit{to break one's heart} or \textit{to turn off}, which have been rarely modeled. By three editions of the PARSEME shared tasks with the corpora \cite{savary-etal-2017-parseme,ramisch-etal-2018-edition,ramisch-etal-2020-edition,savary-etal-2023-parseme}, \textsc{MwE} community is focusing on how to specialized models for robust \textsc{VMwE} identification.

\paragraph{In-Context Learning (ICL)}
Recently, many studies have shown that in-context learning is successful, allowing LLMs to adapt to a task given merely a few input-output pairs in the prompt without any parameter updates \cite{chen2024parallel}, enabling LLMs to ``learn by analogy'' and perform complex tasks. In particular, many conventional  tasks, which are performed by specialized models, start to show promising futures with LLMs in generative paradigm, such as named entity recognition \cite{zhou2023universalner}, information extraction \cite{agrawal2022large} and relation extraction \cite{wadhwa-etal-2023-revisiting}. In the lexical semantics field, \cite{coil-shwartz-2023-chocolate} used the GPT-3 model \cite{brown2020language} with ten-shot settings for interpreting noun compounds. \cite{chakrabarty2022s} also leveraged the GPT-3 model to interpret non-compositional figurative phrases, including idioms and similes. \cite{zhang2024prolex} proposed an ICL-based method for lexical substitution with better language proficiency.

\begin{table*}
\vspace{-0.5em}
\centering
\tiny
\setlength\extrarowheight{2pt}
\begin{tabularx}{0.94\linewidth}{p{1.8cm}p{2.1cm}>{\centering\arraybackslash}p{1.2cm}>{\centering\arraybackslash}p{2.2cm}p{1.1cm}>{\centering\arraybackslash}p{1.9cm}>{\centering\arraybackslash}c}
\toprule
\textbf{Task} & \textbf{Data Source} & \textbf{Input $(\mathcal{I})$} & \textbf{Output $(\mathcal{O})$} & \textbf{Metrics} & \textbf{\# Train / Dev / Test} & \textbf{Phrase Type} \\
\midrule
IE Detection & \cite{tayyar-madabushi-etal-2021-astitchinlanguagemodels-dataset} & $\mathcal{P} \oplus \mathcal{S} \oplus \mathcal{IE}$ & Choice from \textit{Options} & \textsc{Acc} & 140 / 23 / 110 & \textsc{Idiomacity} \\
\midrule
IE Extraction & \cite{Tedeschi2022ID10MII} & $\mathcal{P} \oplus \mathcal{S}$ & Extracted $\mathcal{IE}$ & $\textsc{Acc}_s$ & 33,757 / 3,724 / 447 & \textsc{Idiomacity} \\
\midrule
IE Interpretation & \cite{zhou-etal-2021-pie,chakrabarty2022s} & $\mathcal{P} \oplus \mathcal{S} \oplus \mathcal{IE}$ & Interpretation of $\mathcal{IE}$ & \vspace{-1em}\makecell[l]{\textsc{Rouge-L},\\\textsc{Bert-Score}} & 408 / 81 / 427 & \textsc{Idiomacity} \\
\midrule
NC Compositionality & \cite{garcia-etal-2021-assessing} & $\mathcal{P} \oplus \mathcal{S} \oplus \mathcal{NC}$ & Choice from \textit{Options} & $\textsc{Acc}$ & 118 / 19 / 100 & \textsc{Noun Compound} \\
\midrule
NC Extraction & \cite{garcia-etal-2021-assessing,kolluru-etal-2022-covid} & $\mathcal{P} \oplus \mathcal{S}$ & Extracted $\mathcal{NC}$ & $\textsc{Acc}_s$ & 9,722 / 1,416 / 720 & \textsc{Noun Compound} \\
\midrule
NC Interpretation & \cite{coil-shwartz-2023-chocolate} & $\mathcal{P} \oplus \mathcal{S} \oplus \mathcal{NC}$ & Interpretation of $\mathcal{NC}$ & \vspace{-1.2em}\makecell[l]{\textsc{Rouge-L},\\\textsc{Bert-Score}} & 160 / 28 / 110 & \textsc{Noun Compound} \\
\midrule
LC Categorization  & \cite{espinosa-anke-etal-2021-evaluating} & $\mathcal{P} \oplus \mathcal{T} \oplus \mathcal{S}$ & \vspace{-0.9em}Choice from \textit{Options} & $\textsc{Acc}$ & \vspace{-1.4em}64,851 /  13,856 / 13,905 (sampled 320) & \textsc{Collocation} \\
\midrule
LC Extraction & \cite{fisas-etal-2020-collfren} & $\mathcal{P} \oplus \mathcal{T} \oplus \mathcal{S}$ & Extracted $\mathcal{LC}$ & $\textsc{Acc}_s$ & 1,600 / 160 / 320 & \textsc{Collocation} \\
\midrule
LC Interpretation & \vspace{-1.3em}\cite{espinosa-anke-etal-2019-collocation,espinosa-anke-etal-2021-evaluating} & $\mathcal{P}  \oplus \mathcal{S} \oplus \mathcal{LC}$ & Interpretation of $\mathcal{LC}$ & \vspace{-1.2em}\makecell[l]{\textsc{Rouge-L},\\\textsc{Bert-Score}} & 512 / 82 / 320 & \textsc{Collocation} \\
\midrule
\textsc{VMwE} Extraction & \cite{savary-etal-2023-parseme} & $\mathcal{P} \oplus \mathcal{S}$ & Extracted $\mathcal{VC}$ & $\textsc{Acc}_s$ & 254 / 156 / 475 & \textsc{Verbal MwE} \\
\bottomrule
\end{tabularx}
\caption{A summary of the dataset statistics in \ourbenchmark. $\mathcal{P}$ refers to the prompt template of task instruction, $\mathcal{S}$ represents a specified context or sentence. $\mathcal{T}$ signifies the provided narrative of semantic taxonomy in the prompt. $\mathcal{IE}$ indicates an idiomatic expression, $\mathcal{LC}$ represents a lexical collocation, and $\mathcal{NC}$ stands for a noun compound.}
\label{tab:task-list}
\vspace{-1.2em}
\end{table*}

\section{\ourbenchmark: Semantic Phrase Processing Benchmark}\label{sec:lexbench}

\ourbenchmark\ aims to provide a simple, robust evaluation suites for assessing the level of LMs across diverse semantic phrase processing tasks, drawing on existing data but making substantial changes and supplements to create a uniform framework. 

We formalize the problem of semantic phrase processing as follows: In light of the input prompt template $\mathcal{P}$ and a semantic phrase together with their associated context $\mathcal{S}$, the models are tasked with generating an output denoted as $\mathcal{O}$. For example, in extraction tasks, the model input would be~~$\mathcal{I} :=\mathcal{P} \oplus \mathcal{S}$, and $\mathcal{O}$ denotes the extracted phrase followed the task instruction described in $\mathcal{P}$. The setting of $(\mathcal{P}$, $\mathcal{S}$, $\mathcal{O})$ for each task is detailed in the Table \ref{tab:task-list}. Each dataset $\mathcal{D}:=\{(p_i, s_i, o_i)\}_{i=1}^{N}$ consists of $N$ examples, each containing a prompt $p_i$, a semantic phrase with its context $s_i$, and a gold label $o_i$. As a result, we construct a comprehensive evaluation framework including ten tasks with ten datasets and five auto metrics. Each task is described in the following sub-sections.

\subsection{Idiomatic Expression Detection$_\textsc{ (IED)}$}\label{subsec:idiom-detection}

\paragraph{Task Definition.}
The goal of task is to determine, through the form of multiple choice question (MCQ), whether a given $\mathcal{S}$ containing an $\mathcal{IE}$ aligns with one of the paraphrases or description provided for that idiomatic expression.
\vspace{0ex}
\paragraph{Data.}
We use the test split of dataset from \cite{tayyar-madabushi-etal-2021-astitchinlanguagemodels-dataset}, which consists of 1,688 instances with different interpretation to 273 idioms. We cluster the instances together that has the same $\mathcal{IE}$ and context, and formalize them to the form of multi-choice question with the four options: ``A'', ``B'', ``C'', or ``D'' as the answer. As a result, we obtained a total of 273 samples of MCQ.
\vspace{0ex}
\paragraph{Metric.}
Exact match (EM). For the models that can access the log-likelihood of the next predicted token, we compute by likelihood of prediction directly.
For the models that are disabled to retrieve conditional likelihood, we parse each $\mathcal{O}$ with predefined heuristic rules, then compute by exact-matched patterns compared to the gold standards.

\subsection{Idiomatic Expression Extraction$_\textsc{ (IEE)}$}\label{subsec:Idiom-Extraction}

\paragraph{Task Definition.}Given an idiom $\mathcal{IE}$ in a sentential context $\mathcal{S}$, the goal is to extract the idiom in context. The extraction procedure can be cast to the text span extraction task or identification task like named entity recognition \cite{wang2023gpt}. 
\paragraph{Data.}Initially, we use the English subset of idiom data from ID10M \cite{Tedeschi2022ID10MII} as the data source. In the test split, we filter to keep the instances with only one idiom item. We deduplicate the examples to ensure there are no existing overlapping items, which aim at detecting the robustness of the models in handling diverse idioms. 
\paragraph{Metric.}In tradition, \textit{token-level F1} is often employed by people since idiom extraction is cast to sequence labeling tasks in identification format. But we determine to use the phrase-level evaluation metrics, namely \textit{sequence accuracy} ($\textsc{Acc}_s$), to adapt the generative extraction style of LLMs. 

\subsection{Idiomatic Expression Interpretation$_\textsc{ (IEI)}$}\label{subsec:Idiom-Interpretation}

\paragraph{Task Definition.}The aim of idiom interpretation is to paraphrase an idiom $\mathcal{IE}$ in a given context $\mathcal{S}$ to reveal its non-compositional meaning. \vspace{0ex}
\paragraph{Data.}We use the annotated idiom paraphrase data from \cite{zhou-etal-2021-pie} and \cite{chakrabarty2022s}. We retained three properties for each sample: (i) idiom, (ii) context, and (iii) annotation meaning while also deduplicating samples from the two sources based on the unique idioms. \vspace{0ex} 
\paragraph{Metric.}ROUGE-L \cite{lin-2004-rouge}, BERT Score \cite{zhang2019bertscore} and PPL \cite{jelinek1977perplexity}.

\subsection{Noun Compound Compositionality$_\textsc{ (NCC)}$}\label{subsec:NC-Compositionality}

\paragraph{Task Definition.}The target of $\mathcal{NC}$ Compositionality is to detect and judge the degree of Compositionality of the noun compound (i.e., the extent to which the meaning of the $\mathcal{NC}$ can be construed as a combination of its constituents.)\vspace{0ex}
\paragraph{Data.}We use the data of NCTTI \cite{garcia-etal-2021-assessing} and convert each phrase-level instance to the same MCQ problem format as IED (See \S\ref{subsec:idiom-detection}). In the end, we compiled a total of 237 examples. \vspace{0ex}
\paragraph{Metric.} Similar to IED, we use exact match to measure the tested systems in the MCQ problem.

\begin{table}
\centering
\scriptsize
\begin{tabular}{l@{\hspace{-2.5em}}c@{\hspace{-1.4em}}r}
\toprule
\textbf{\underline{L}exical \underline{F}unction} & \textbf{Example} & \textbf{Semantic Relation} \\
\midrule
Magn & Magn(\textit{rain}) = \textit{heavy} & ``intense'', ``strong'' \\
AntiMagn & AntiMagn(\textit{accent}) = \textit{slight} & ``little'', ``weak'' \\
\midrule
Ver & Ver(\textit{message}) = \textit{clear} & ``real'', ``genuine''\\
AntiVer & AntiVer(\textit{accusation}) = \textit{groundless} & ``non-genuine'' \\
\midrule
Bon & Bon(\textit{bread}) = \textit{fresh} & ``positive'' \\
AntiBon & AntiBon(\textit{advantage}) = \textit{undue} & ``negative'' \\
\midrule
Son & Son({alarm clock}) = \textit{ring(s)} & ``sound'', ``voice'' \\
\midrule
Oper1 & Oper1(\textit{advice}) = \textit{give} & ``perform'' \\
\bottomrule
\end{tabular}
\caption{Partial semantic relations (i.e., lexical functions) involved in this paper, with their exemplars. More relations in LF can be referred to the Table \ref{tab:lexfunc-with-semantic-gloss}.}
\label{tab:collocation-semantic-relations}
\vspace{-1.5em}
\end{table}

\subsection{Noun Compound Extraction$_\textsc{ (NCE)}$}\label{subsec:NC-Extraction}

\paragraph{Task Definition.}Given a sentence $\mathcal{S}$ that includes only one noun compound $\mathcal{NC}$, the target is to extract the $\mathcal{NC}$ correctly in context, including the constituents of \textit{head word} and \textit{modifier word}.\vspace{0ex}
\paragraph{Data.} We compiled and sampled 720 examples from the dataset \textsc{ProNCI} \cite{kolluru-etal-2022-covid}.\vspace{0ex}
\paragraph{Metric.}Sequence-level accuracy ($\textsc{Acc}_s$).

\subsection{Noun Compound Interpretation$_\textsc{ (NCI)}$}\label{subsec:NC-Interpretation}

\paragraph{Task Definition.}Given a noun compound $\mathcal{NC}$, the target of task is to interpret this $\mathcal{NC}$ with its literal meaning concluded from the constituents.\vspace{0ex}
\paragraph{Data.} We use the revised dataset of \cite{coil-shwartz-2023-chocolate} based on \cite{hendrickx-etal-2013-semeval}, which consists of $298$ noun compounds with $11,730$ annotated paraphrases in total.\vspace{0ex}
\paragraph{Metric.}ROUGE-L, BERT Score, and PPL.

\subsection{Lexical Collocation Categorization$_\textsc{ (LCC)}$}\label{subsec:Collocation-Categorization}

\paragraph{Task Definition.}Given a sentence $\mathcal{S}$ containing a collocation $\mathcal{LC}$, the target is to correctly classify the collocation conditioned with its context $\mathcal{S}$ to the proper category, according to the defined semantic relation taxonomy (cf. Table \ref{tab:collocation-semantic-relations}).
\paragraph{Data.}We reuse the train and validation splits from \cite{espinosa-anke-etal-2021-evaluating}, but we downsample 40 instances from each of the eight categories so that it is feasible to use accuracy to measure models' performance across multiple classes.
\paragraph{Metric.} Accuracy\footnote{We use accuracy for sequence classification models and compute the same metric but with exact matching for LLMs.} ($\textsc{Acc}$).

\subsection{Lexical Collocation Extraction$_\textsc{ (LCE)}$}\label{subsec:Collocation-Extraction}

\paragraph{Task Definition.}Collocation extraction refers to the process of identifying \textit{base word} and \textit{collocate word} together from a given sentence $\mathcal{S}$. This task involves recognizing a textual phrase, optionally in a given specific semantic relation (cf. Table \ref{tab:collocation-semantic-relations}).
\paragraph{Data.}We construct the test set sampled from the English part of collocation identification data of \cite{fisas-etal-2020-collfren,espinosa-anke-etal-2022-multilingual} but for extraction task usage. And the datasets of training and validation are reused from them.
\paragraph{Metric.} Sequence-level accuracy ($\textsc{Acc}_s$).

\subsection{Lexical Collocation Interpretation$_\textsc{ (LCI)}$}\label{subsec:Collocation-Interpretation}

\paragraph{Task Definition.} Similar to idiom interpretation, the task aims to examine the model $\mathcal{M}$ for interpreting the contextualized meanings according to the given pair-wise $\mathcal{LC}$-$\mathcal{S}$ examples.
\paragraph{Data.}We use an in-context collocation dataset, the expanded \textsc{LexFunc} \cite{espinosa-anke-etal-2021-evaluating}, as our initiation and sample 40 data points for each relation category (cf. Table \ref{tab:collocation-semantic-relations}) from it. Then, each collocation is annotated with five different paraphrases by three judges, and the annotation guidelines are shown in the Appendix \S\ref{sec:annotation-guideline}. 
\paragraph{Metric.} ROUGE-L, BERT Score, and PPL.

\subsection{Verbal \textsc{MwE} Extraction$_\textsc{ (VMwE)}$}

\paragraph{Task Definition.}
Given a sentence $\mathcal{S}$ that includes one of the verbal constructions from $\mathcal{VPC}$, $\mathcal{LVC}$, and $\mathcal{VID}$, the goal of task is to extract the existing  construction correctly from its context.\vspace{0ex}
\paragraph{Data.}We use a well-known annotated \textsc{VMwE} dataset, \text{PARSEME-corpus-release-1.3}
\cite{savary-etal-2023-parseme}, and we process the data to make sure each data point only contains one $\mathcal{VC}$ in context.\vspace{0ex}
\paragraph{Metric.}
Sequence-level accuracy ($\textsc{Acc}_s$).

\section{Experimental Setup}
\label{sec:length}

\paragraph{Datasets.}
We introduced our curated datasets illustrated in \S\ref{sec:lexbench} and \S\ref{sec:appendix-additional-experiment-details}. Each subset has a specific ratio of train, validation, and test splits. Rate limitations on using the API-based LLMs prevented us from studying the full set performance, so we selected random samples from the initial test set. Few-shot examples are sampled from the original training set of tasks and prompted to the models.

\addtocounter{figure}{-1}

\begin{figure*}[t]
    \centering
    \subfigure{
	    \begin{minipage}{0.313\linewidth}
	    \includegraphics[width=\linewidth]{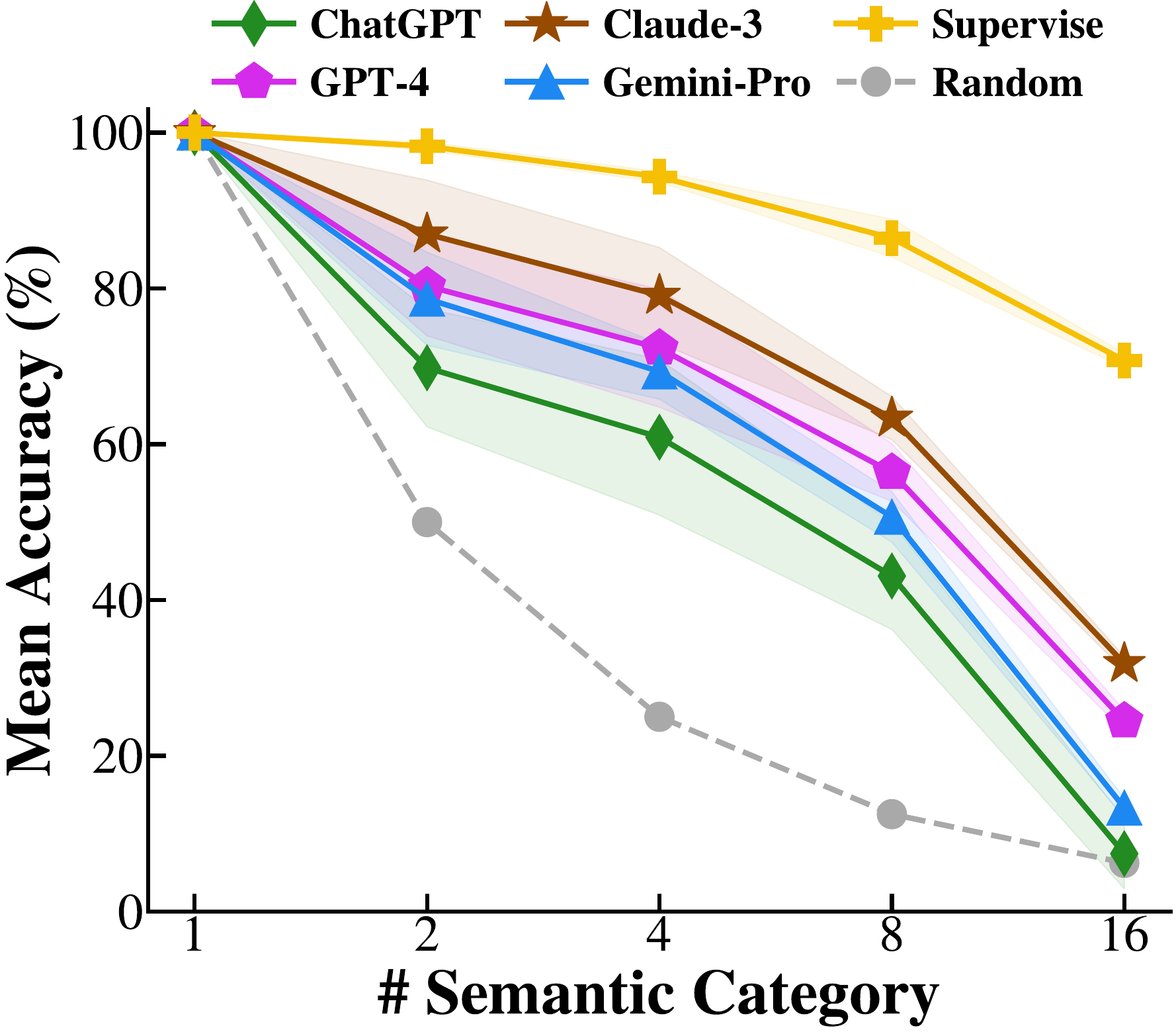}
	    \vspace{-1.1em}
	    \caption{Each model is run with zero-shot prompting in the semantic relation classification with category scaling. Mean accuracy scores $(\%)$ of different models are average over runs in three sampled sets. For comparative reasons, we also plotted the level of random baseline.}
	    \label{fig:lcc-scaling-comparison}
	    \end{minipage} 
    }
    \hfill\hspace{0em}
    \subfigure{
	    \begin{minipage}{0.313\linewidth}
	    \vspace{.2em}
	    \includegraphics[width=\linewidth]{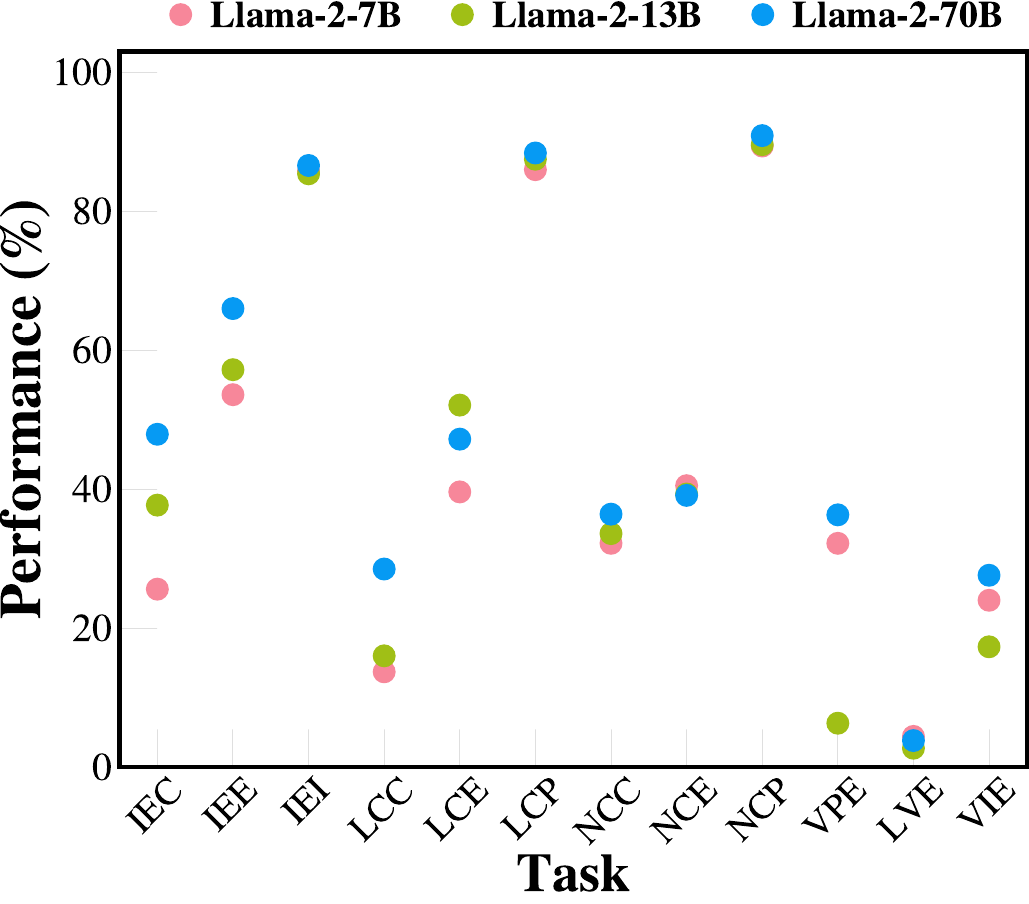}
	    \vspace{-1.1em}
	    \caption{Impact of Models' scale (7B, 13B, and 70B of Llama 2) on the performance in \ourbenchmark\ for all twelve tasks.  Note that values of different task formats should not be compared since the y-axis delineates the performance of each task with the corresponding metric.}
	    \label{fig:llama-scaling-comparison}
	    \end{minipage} 
    }
    \hfill\hspace{0em}
    \subfigure{
	    \begin{minipage}{0.313\linewidth}
	    \includegraphics[width=\linewidth]{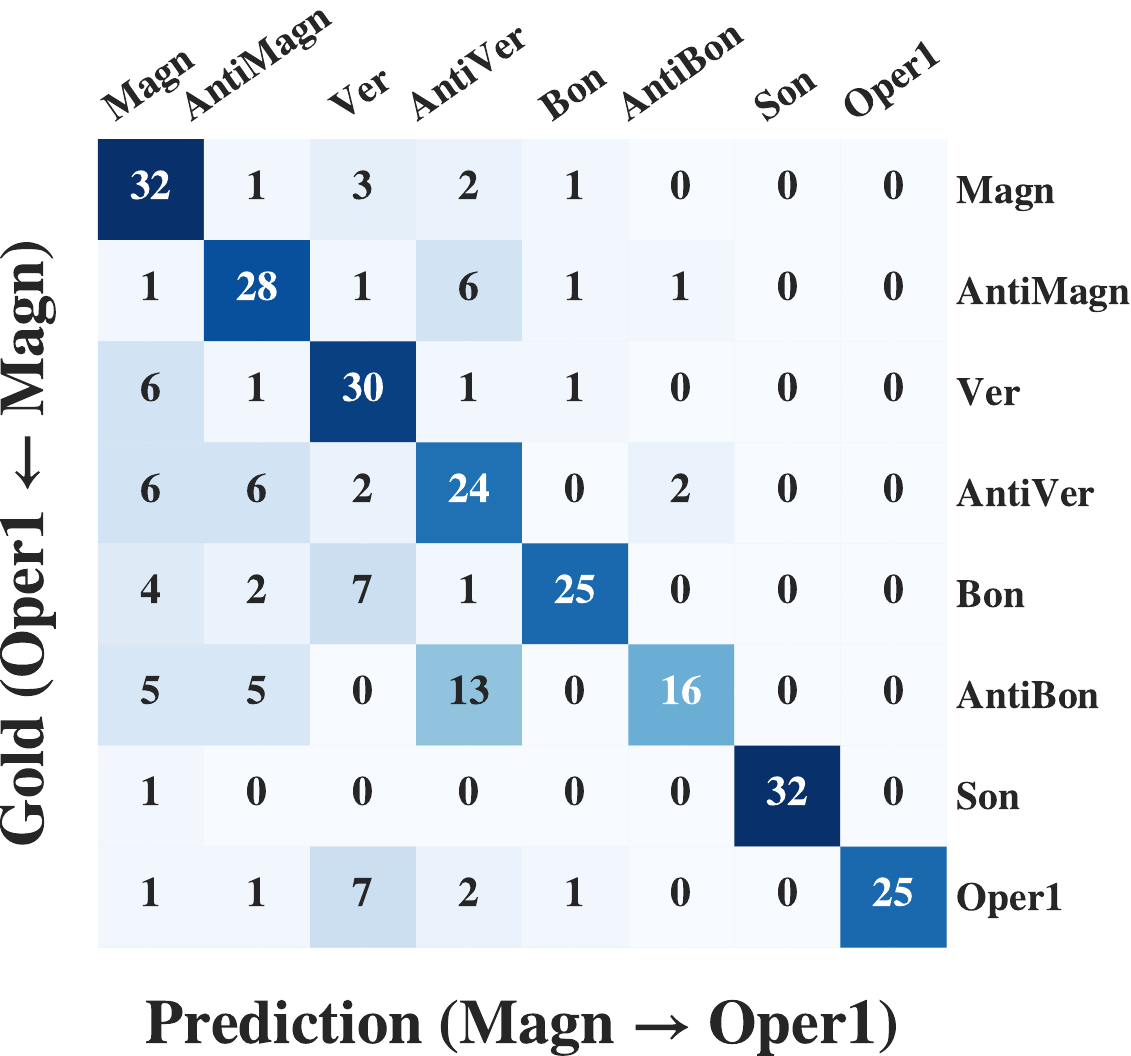}
	    \vspace{-1.1em}
	    \caption{Confusion matrix for the best-performing model with ICL (Claude-3 in 5-shot setting) in categorizing eight semantic relations described by lexical functions (cf. Table \ref{tab:collocation-semantic-relations}). The x-axis denotes the prediction results, and the y-axis represents the gold standards.}
	    \label{fig:lcc-eight-classes-cm}
	    \end{minipage}
    }
    \vspace{-1.3em}
\end{figure*}

\paragraph{Models.}

As our experimental baselines, we adopt fifteen currently popular NLP systems across different architectures and model scales, with strong semantic understanding ability. Open-source models include \text{BERT-base / large} \cite{devlin2019bert}, \text{T5-base / large} \cite{raffel2020exploring}, \text{Llama-2-7B / 13B / 70B-chat} \cite{touvron2023llama2}, \text{DeepSeek-7B-chat} \cite{bi2024deepseek}, \text{Mistral-7B-instruct-v0.1} \cite{jiang2023mistral} and \text{Mixtral-8x7B-instruct-v0.1} \cite{jiang2024mixtral}. While the selected proprietary models include \text{GPT-3.5-Turbo-0613} (ChatGPT) \cite{openai_chatgpt}, \text{GPT-4-1106-Preview} (GPT-4) \cite{openai2023gpt4}, and \text{Claude-instant-1} \cite{anthropic-model-card}, \text{Claude-3-Opus-20240229} (Claude-3) \cite{TheC3}, as well as \text{Gemini-1.0-Pro} (Gemini-Pro) \cite{geminiteam2023gemini} provided by Google. Table \ref{tab:models} lists all the details of the LMs we considered.

\begin{table*}
\centering
\tiny
\setlength\tabcolsep{4pt}
\setlength{\extrarowheight}{2pt}
\begin{tabular}{lcccccccccccccccc}
\toprule
\multirowcell{2.7}{\textbf{{\scriptsize{M}}ODEL}} & \multicolumn{3}{c}{\textbf{{\scriptsize{I}}DIOM}} & \phantom{c} & \multicolumn{3}{c}{\textbf{{\scriptsize{C}}OLLOCATION}} & \phantom{c} & \multicolumn{3}{c}{\textbf{{\scriptsize{N}}OUN {\scriptsize{C}}OMPOUND}} & \phantom{c} & \multicolumn{3}{c}{\textbf{{\scriptsize{VM}}W{\scriptsize{E}}}} & \phantom{c} \\
\cmidrule{2-4}\cmidrule{6-8}\cmidrule{10-12}\cmidrule{14-16}
& \textbf{IED} & \textbf{IEE} & \textbf{IEI} && \textbf{LCC} & \textbf{LCE} & \textbf{LCI} && \textbf{NCC} & \textbf{NCE} & \textbf{NCI} && \textbf{VPE} & \textbf{LVE} & \textbf{VIE} & \\
\midrule
\makecell{\textbf{{\scriptsize{M}}ETRIC~\raisebox{0.23ex}{$(\%)$}}} & $\textsc{Acc}$ & \hspace{0.05cm}$\textsc{Acc}_s$ & B-S && \textsc{Acc} & \hspace{0.05cm}$\textsc{Acc}_s$ & B-S && $\textsc{Acc}$ & \hspace{0.05cm}$\textsc{Acc}_s$ & B-S && \hspace{0.05cm}$\textsc{Acc}_s$ & \hspace{0.05cm}$\textsc{Acc}_s$ & \hspace{0.05cm}$\textsc{Acc}_s$ \\
\midrule
\underline{\textbf{{{\scriptsize{H}}UMAN}}} & 71.0 & \uline{\textbf{87.0}} & 87.6 && 47.0 & 50.0 & 86.8 && \uline{\textbf{71.0}} & 73.0 & 80.3 && \uline{\textbf{85.0}} & \uline{\textbf{55.0}} & \uline{\textbf{78.0}} \\
\midrule
\vspace{0.07em}\underline{\textbf{{\scriptsize{S}}UPERVISED {\scriptsize{M}}ETHODS}} & & & & & & & & & & & & & & & & \\
\hlcelll \textsc{BERT}$_\textsc{B}$: \textit{fine-tuned} & 85.0 & 66.8 & - && 78.8 & 63.1 & - && 53.6 & 68.5 & - && 68.7 & \textcolor{ggreen}{\textbf{52.2}} & 36.1 & \\
\hlcelll \textsc{BERT}$_\textsc{L}$: \textit{fine-tuned} & \textcolor{ggreen}{\textbf{85.1}} & 67.2 & - && \textcolor{ggreen}{\textbf{82.6}} & 63.8 & - && 51.5 & 69.1 & - && 74.1 & 41.7 & 34.2 & \\
\hlcelll \textsc{T5}$_\textsc{B}$: \textit{fine-tuned} & - & - & 86.8 && - & - & 87.2  && - & - & 89.7 && - & - & - & \\
\hlcelll \textsc{T5}$_\textsc{L}$: \textit{fine-tuned} & - & - & 87.1 && - & - & 87.7  && - & - & 89.8 && - & - & - & \\
\midrule
\vspace{0.07em}\underline{\textbf{{\scriptsize{I}}NFERENCE-ONLY {\scriptsize{M}}ETHODS}} & & & & & & & & & & & & & \\
\hlcellll DeepSeek-7B: \textit{zero-shot} & 37.0 & 52.3 & \textcolor{rred}{\textbf{80.8}} & & \textcolor{rred}{\textbf{13.4}} & \textcolor{rred}{\textbf{25.9}} & 86.1 & & 30.9 & \textcolor{rred}{\textbf{29.7}} & 89.4 && 28.1 & 7.6 & 20.3 \\
\hlcellll Mistral-7B: \textit{zero-shot} & 47.2 & 58.8 & 85.7 & & 34.7 & 43.2 & \textcolor{rred}{\textbf{85.1}} && \textcolor{rred}{\textbf{29.6}} & 46.2 & \textcolor{rred}{\textbf{87.6}} && 6.4 & 3.9 & \textcolor{rred}{\textbf{12.5}} \\
\hlcellll Mixtral-8x7B: \textit{zero-shot} & 48.3 & 68.2 & 86.6 && 36.1 & 50.8 & 85.2 && 36.0 & 64.0 & 91.4 && 26.3 & 6.1 & 24.7 \\
\hlcellll Llama-2-7B: \textit{zero-shot} & \textcolor{rred}{\textbf{25.6}} & 53.6 & 85.8 & & 13.7 & 39.6 & 86.0 & & 32.2 & 40.5 & 89.4 && 32.2 & 4.4 & 24.0 \\
\hlcellll Llama-2-13B: \textit{zero-shot} & 37.7 & 57.2 & 85.4 && 16.0 & 52.1 & 87.5 && 32.6 & 39.3 & 89.6 && \textcolor{rred}{\textbf{6.3}} & \textcolor{rred}{\textbf{2.7}} & 17.3 \\
\hlcellll Llama-2-70B: \textit{zero-shot} & 47.9 & 66.0 & 86.6 && 28.5 & 47.2 & 88.4 && 36.4 & 39.1 & 90.9 && 36.3 & 3.8 & 27.6 \\
\hdashline
\hlcell GPT-3.5-Turbo: \textit{zero-shot} & 40.6 & 68.9 & 85.6 && 32.4 & 53.4 & 88.9 & & 41.9 & 67.2 & 91.4 && 60.0 & 7.7 & 42.8 & \\
\hlcell \hspace{0.88cm}$\hookrightarrow \text{+}$ \textit{three-shot} & 45.4 & 67.3 & 88.2 & & 36.3 & 69.5 & 92.4 & & 43.6 & 72.7 & 96.5 & & 53.6 & 10.0 & 30.4 & \\
\hlcell \hspace{0.88cm}$\hookrightarrow \text{+}$ \textit{five-shot} & 46.5 & 67.7 & 88.3 & & 40.9 & 71.1 & 92.4 & & 39.1 & 69.1 & 96.9 && 58.9 & 15.0 & 31.4 & \\
\hlcell GPT-4.0-Turbo: \textit{zero-shot} & 66.3 & 75.1 & 86.5 & & 53.4 & 70.1 & 89.4 & & 53.4 & 75.4 & 89.9 && 61.5 & 7.7 & 42.8 & \\
\hlcell \hspace{0.88cm}$\hookrightarrow \text{+}$ \textit{three-shot} & 70.3 & 77.1 & 88.1 & & 60.0 & 77.7 & \textcolor{ggreen}{\textbf{92.9}} & & 56.3 & 83.6 & 94.8 && 75.8 & 16.1 & 43.8 & \\
\hlcell \hspace{0.88cm}$\hookrightarrow \text{+}$ \textit{five-shot} & 72.8 & 72.7 & \textcolor{ggreen}{\textbf{88.4}} & & 58.1 & \textcolor{ggreen}{\textbf{77.8}} & 92.7 & & 58.6 & \textcolor{ggreen}{\textbf{85.4}} & 95.5 && \textcolor{ggreen}{\textbf{77.8}} & 13.3 & \textcolor{ggreen}{\textbf{48.5}} & \\
\hlcell Claude-Instant-1: \textit{zero-shot} & 51.2 & 72.2 & 85.7 && 40.5 & 42.6 & 89.7 && 43.2 & 50.9 & 91.9 && 59.2 & 11.6 & 39.0 & \\
\hlcell \hspace{1cm}$\hookrightarrow \text{+}$ \textit{three-shot} & 47.9 & 60.8 & 86.5 && 49.8 & 54.7 & 87.0 & & 47.8 & 59.1 & 94.1 && 48.9 & 18.8 & 35.5 & \\
\hlcell \hspace{1cm}$\hookrightarrow \text{+}$ \textit{five-shot} & 52.0 & \textcolor{rred}{\textbf{47.4}} & 87.0 & & 50.1 & 57.7 & 87.1 & & 44.9 & 61.8 & 94.5 && 53.1 & 15.0 & 38.4 & \\
\hlcell Claude-3-Opus: \textit{zero-shot} & 66.3 & 62.8 & 87.1 && 61.3 & 34.7 & 88.5 & & 50.4 & 36.3 & 91.7 && 67.3 & 28.3 & 42.8 & \\
\hlcell \hspace{0.88cm}$\hookrightarrow \text{+}$ \textit{three-shot} & 75.8 & 64.8 & 88.1 && 69.5 & 56.7 & 92.8 && 56.7 & 33.6 & 93.1 && 74.7 & 37.2 & 47.6 \\
\hlcell \hspace{0.88cm}$\hookrightarrow \text{+}$ \textit{five-shot} & 72.8 & 67.1 & 88.2 && 69.8 & 60.0 & 92.8 & & \textcolor{ggreen}{\textbf{63.9}} & 30.9 & 96.0 && 75.7 & 35.5 & 43.2 \\
\hlcell Gemini-Pro: \textit{zero-shot} & 56.0 & \textcolor{ggreen}{\textbf{77.8}} & 86.9 & & 48.5 & 51.8 & 89.5 && 38.5 & 59.0 & 91.8 & & 43.8 & 6.7 & 43.8 & \\
\hlcell \hspace{0.59cm}$\hookrightarrow \text{+}$ \textit{three-shot} & 49.4 & 76.0 & 88.1 & & 55.5 & 64.9 & 92.6 & & 36.4 & 65.4 & 95.2 & & 51.5 & 13.3 & 42.8 & \\
\hlcell \hspace{0.59cm}$\hookrightarrow \text{+}$ \textit{five-shot} & 50.9 & 76.5 & 88.3 & & 55.7 & 66.8 & 92.3 & & 34.3 & 74.5 & \textcolor{ggreen}{\textbf{97.7}} & & 52.6 & 11.6 & 44.7 & \\
\bottomrule
\end{tabular}
\caption{Major experimental results in \ourbenchmark. ``-'' denotes the model that is unavailable or inappropriate for the task. \uline{\textbf{Digits}} denotes the result of humans is better than all models. \colorbox{sred}{Light Pink} text delineates the baselines with supervised fine-tuning. \colorbox{sgreen}{Light Green} and \colorbox{sblue}{Light Blue} parts present open-source models and proprietary models.}
\vspace{-4ex}
\label{tab:major-exp-results}
\end{table*}

\paragraph{Metrics.}The metrics for each task are shown in Table \ref{tab:task-list} and illustrated in \S\ref{sec:lexbench}. Since the benchmark involves classification and generation tasks, the metrics use either prediction-based ones like accuracy (\textsc{Acc}), exact match (EM)\footnote{We use heuristic rules to parse the responses of each LLM according to its characteristics of feedback.}, or ngram-based ROUGE-L (R-L), and semantic similarity-based BERT Score (B-S), depending on the performing task (\S\ref{sec:lexbench}). While existing works commonly utilize precision, recall, and F1 to evaluate the effect of systems in \textsc{VMwE} identification \cite{pasquer-etal-2020-verbal,tanner-hoffman-2023-mwe}, for the sake of comparability (owing to the characteristics of LLM-based generative extraction), we adopt phrase-level accuracy ($\textsc{Acc}_s$) as the metric.

\paragraph{Implementation Details.} 

We probe the zero-shot and few-shot (three- and five-shot) performance for the inference-only models. Evaluation on few-shot prompting with open-source models is left for future work. We utilized temperature with $\tau = 0$ on decoding parameters and used top-p decoding \cite{holtzman2019curious} with $p = 1.0$ for all the models. The inference is deployed by vLLM \cite{kwon2023efficient}. More experimental details can be found in the Appendix \S{\ref{sec:additional-exp-details}}.

\section{Results}
We provide a summary table of major experiment results in Table \ref{tab:major-exp-results} and additional results in Table \ref{tab:perf-interpretation}.

\subsection{Benchmarking Results}

\paragraph{Main results}
Table \ref{tab:major-exp-results} and Figure \ref{fig:Performance-SpiderFigure} showcase the performance in percentage (\%) of
various evaluated models in \ourbenchmark. We summarize the key findings from the experimental results: (i) Significantly, GPT-4 $(\texttt{gpt-4-1106-preview})$ demonstrates comprehensive and superior performance compared to other models across various phrase types in the categorization, extraction, and interpretation tasks exhibiting notably higher average scores and achieving the top tier in six out of twelve sub tasks. (ii) In the interpretation tasks, inference-based models perform relatively well with the supervised baselines. In other words, most proprietary models outperform in these tasks (i.e., IEI, LCI, and NCI). On the other hand, larger models still behave poorly than finetuned models in multi-class categorization and extraction tasks, such as IED and IEE. (iii) It is evident that in most of the tasks, the current state-of-the-art open-source models (e.g., Mixtral-8x7B-inst and Llama-70B-chat) still lag behind several proprietary models in roughly the same level of model size, indicating that there remains a considerable performance gap between the two, in terms of the domain of semantic phrase processing.

\paragraph{Differences between Task Types}
Comparing categorization and extraction tasks, we observe a significant variance in the distribution of performance to the same model. For instance, zero-shot Gemini-Pro yields $56.0\%$ in IED but $77.8\%$ in IEE. When comparing the extraction and interpretation tasks, each model still displays distinct differences in performance of the exact phrase type (e.g., zero-shot GPT-3.5-Turbo is $68.9\%$ in IEE but with a much higher score of $85.6\%$ in IEI). Similarly, the same gap occurs in the classification and interpretation tasks. For example, zero-shot GPT-3.5-Turbo yields $40.6\%$ in IED but $85.6\%$ in IEI. Due to different metrics, the scores in the same type of phrase indicate that the results may not be compared directly between different task formats; GPT-4 exhibits balanced scores in the tasks. It unveils larger models that equip more balanced performance across various tasks. 

\paragraph{Relationship with Scaling Law.} As shown in Table \ref{tab:major-exp-results} and Figure \ref{fig:llama-scaling-comparison}, the ability of Llama models increases steadily as the number of model parameters scales up. This finding strongly implies that the overall capacity of LLMs in semantic phrase processing is determined by the Scaling Law \cite{kaplan2020scaling}. The model scale is critical for Llama to perform semantic phrase processing. For instance, the significant accuracy increase with 7B, 13B, and 70B model scales in IED is $25.6\% \rightarrow 37.7\% \rightarrow 47.9\%$. However, as reported in Table \ref{tab:perf-interpretation}, it does not exhibit a significant gap between large and smaller models in semantic similarity-based measurement for the three interpretation tasks. For example, zero-shot Llama-70B ($86.6_\textit{B-S}$) is only $1.2\%$ higher than zero-shot Llama-13B ($85.4_\textit{B-S}$) and $1.0\%$ higher than zero-shot Llama-7B ($85.8_\textit{B-S}$) in IEI. In contrast, Llama-2-70B shows larger superiority in IEE, compared to Llama-2-13B ($66.0_\textsc{Acc}$ vs. $57.2_\textsc{Acc}$), as well as Llama-2-7B ($66.0_\textsc{Acc}$ vs. $53.6_\textsc{Acc}$). The results may be attributed to the fact that the insufficient sensitivity of similarity-based metrics which also is reported in \cite{Bai2023LongBenchAB}.

\paragraph{Effect of In-Context Learning.}The results of experiments (cf. Table \ref{tab:major-exp-results} and Table \ref{tab:perf-interpretation}) highlight the effectiveness of ICL. We observed that many tasks had benefits, but some of the others had decayed in performance. As the number of demonstration examples increases from 0 to 3 and 5 in the interpretation tasks, the BERT Score of the three interpretation tasks increases sufficiently. In contrast, we also find that few-shot demonstrations may cause performance regression in some tasks. For instance, in IEE, besides Claude-3's consistent gains under in-context learning, all other models exhibited varying degrees of performance regression as the number of demonstration exemplars increased. Specifically, the accuracy of Claude-instant-1 decreased from $72.2$ in zero-shot to $60.8$ and $47.4$ in 3-shot and 5-shot settings, respectively. This perhaps suggests the importance of exemplars selection, and we left this assumption for future work.

\addtocounter{figure}{+1}

\begin{figure*}[t]
    \centering
    \begin{minipage}{1.0\linewidth}
    \includegraphics[width=\linewidth]{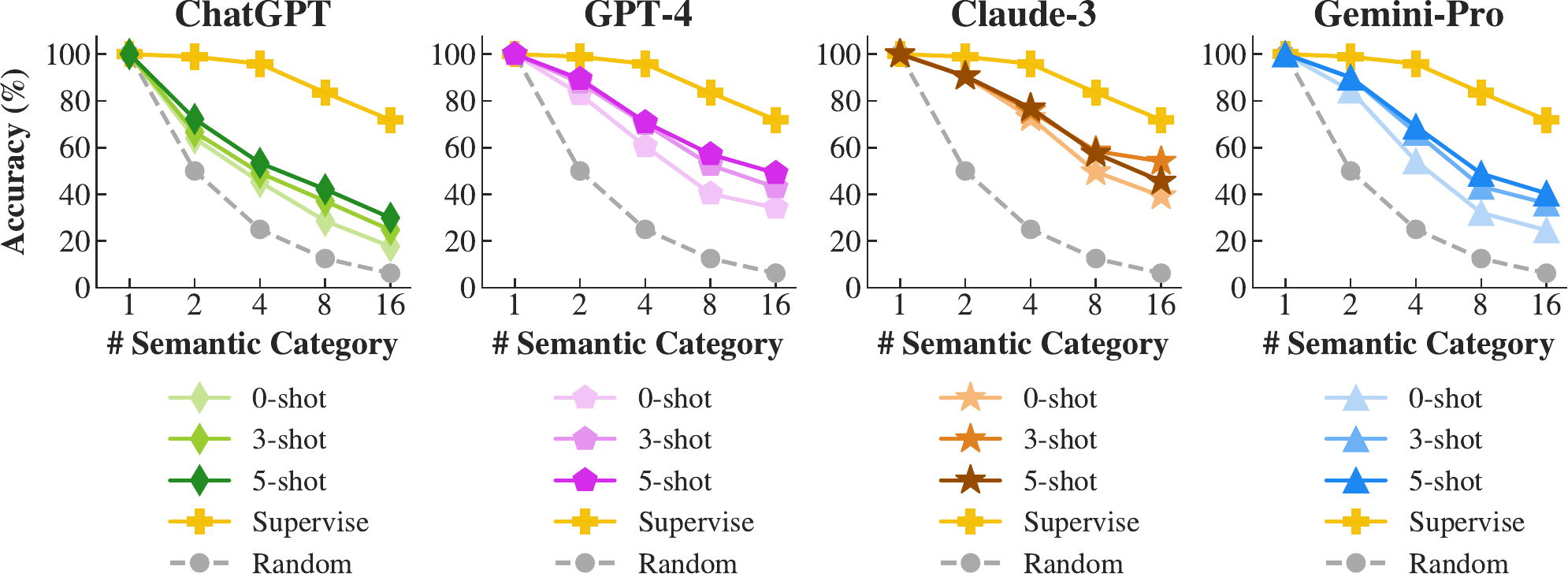}
    \end{minipage}
    \vspace{-0.5em}
    \caption{The ability of semantic relation categorization of $\mathcal{LC}$ with different numbers of in-context exemplars and semantic category scale. The number $n$ of classes is chosen from $N:= \{1, 2, 4, 8, 16\}$. Each model is prompted with the $k$-shot settings, where $k \in \{0, 3, 5\}$, respectively. Accuracy scores are calculated by the mean values based on 30 examples sampled per class from the test split of \cite{espinosa-anke-etal-2021-evaluating}, partial categories $(n \le 8)$ are run with three-class combinations in random selection, finally result in the mean value as the average.}
    \label{fig:lcc-scaling-detail}
    \vspace{-0em}
\end{figure*}

\subsection{Human Performance}
We estimate human performance to provide a more reliable measurement than automatic metrics by employing three graduate students who majored in linguistics to annotate 100 random examples sampled from the test set of each task in \ourbenchmark. We adopt a two-stage approach, wherein an annotator undergoes a brief training phase before advancing to the annotation phase. This process is inspired by the methodology employed by SuperGLUE \cite{sarlin2020superglue}. Our results (cf. Table \ref{tab:major-exp-results},\ref{tab:perf-interpretation}) show that humans only dominate in three out of ten tasks, with models exhibiting superior performance in the rest.\ Therefore, we conclude that the LM's ability to understand semantic phrases has reached parity with humans and exceeded it in most tasks.

\subsection{Semantic Category Scaling with In-Context Learning}
To investigate the semantic understanding capacity of LMs regarding general phrases (i.e., lexical collocation), we further explored the categorization performance under varying numbers of semantic category combinations. In detail, we pick classes of lexical relations from \ourbenchmark\ with at least 40 cases each and assess four strong models' performance on these cases. As a result, figure \ref{fig:lcc-scaling-comparison} reports the performance on the scaling LCC task in \ourbenchmark. In Figure \ref{fig:lcc-scaling-detail}, we demonstrate the accuracy variations of four strong models in $N$-classification tasks, where $N \in \{1, 2, 4, 8, 16\}$. We also run the one-class classification to ablate the impact of the instruction-following capacity of models (per-class results are shown in Table \ref{tab:lcc-scaling-exp-detail} in the Appendix \S\ref{subsec:lcc-exp}). We find that the selected models outperform the corresponding random baseline, respectively. However, not surprisingly, their accuracy all exhibits linear decay as the number of potential target categories increases. Specifically, zero-shot ChatGPT regress nearly to the random baseline $(17.69\%\ \text{vs.}\ 6.25\%)$ in 16-class classification as shown in the Figure \ref{fig:lcc-scaling-comparison}.

\begin{table}[t]
\renewcommand{\arraystretch}{1.1}
\scriptsize
\centering
\begin{tabular}{p{1.55cm}>{\centering\arraybackslash}p{0.8cm}cp{0.1em}c}
\toprule
\multirow{2.5}{*}{\textbf{System}} & \multirow{2.5}{*}{\textbf{Setting}} & \multicolumn{1}{l}{\textbf{w/ \textsc{Oracle}}} && \multicolumn{1}{l}{\textbf{w/o \textsc{Oracle}}} \\ \cmidrule{3-5} && \textsc{Acc}$(\Delta)$ \raisebox{0.2ex}{$\uparrow$} && \textsc{Acc} \raisebox{0.2ex}{$\uparrow$} \\
\midrule
BERT$_\text{B}$ & SFT & - && 52.3 \\
\midrule
\multirow{3}{*}{Gemini-1.0-Pro} & \textit{0-shot} & 35.2 {\color{OliveGreen}{(+5.5)}} & \tiny$\longleftrightarrow$ & 29.7 \\  
 & \textit{3-shot} & 47.0 \color{OliveGreen}{(+11.8)} & \tiny$\longleftrightarrow$ & 35.2 \\ & \textit{5-shot} & \textbf{51.1\color{OliveGreen}{(+15.8)}} & \tiny$\longleftrightarrow$ & 35.3 \\ \midrule
\multirow{3}{*}{GPT-3.5-Turbo} & \textit{0-shot} & 38.3 \color{OliveGreen}{(+2.1)} & \tiny$\longleftrightarrow$ & 36.2 \\  
 & \textit{3-shot} & 45.8 \color{OliveGreen}{(+13.8)} & \tiny$\longleftrightarrow$ & 32.0 \\ & \textit{5-shot} & \textbf{46.3 \color{OliveGreen}{(+10.1)}} & \tiny$\longleftrightarrow$ & 36.2 \\ \midrule
 \multirow{3}{*}{GPT-4-Turbo} & \textit{0-shot} & 40.4 \color{OliveGreen}{(+3.4)} & \tiny$\longleftrightarrow$ & 37.0 \\  
 & \textit{3-shot} & 54.6 \color{OliveGreen}{(+8.4)} & \tiny$\longleftrightarrow$ & 46.2 \\ & \textit{5-shot} & \textbf{55.7 \color{OliveGreen}{(+8.8)}} & \tiny$\longleftrightarrow$ & 46.9 \\
 \midrule
  \multirow{3}{*}{Claude-3-Opus} & \textit{0-shot} & \textbf{60.6 \color{OliveGreen}{(+13.5)}} & \tiny$\longleftrightarrow$ & 47.1 \\  
 & \textit{3-shot} & 59.1\color{OliveGreen}{(+4.6)} & \tiny$\longleftrightarrow$ & 54.5 \\ & \textit{5-shot} & 55.1\color{OliveGreen}{(+1.9)} & \tiny$\longleftrightarrow$ & 53.2 \\
\bottomrule
\end{tabular}
\caption{We report four strong models' mean accuracy of the \textsc{VMwE} extraction in different ICL settings and whether employing the \textsc{Oracle Prompting}. \textsc{Oracle} means providing the specific definition of target and the same category of few-shot examples rather than randomly selected ones in the extraction procedure.}
\label{tab:vmwe-oracle-exp}
\vspace{-1.5em}
\end{table}

\subsection{\textsc{VMwE} Extraction via \textsc{Oracle Prompting}}
As an additional analysis, we explore the strategies of prompting methods with in-context learning in the \textsc{VMwE} extraction of \ourbenchmark. These methods include zero-shot and few-shot (i.e., three- and five-shot) prompting. Furthermore, we propose a simple and effective method, namely the \textsc{Oracle Prompting}, for more accurate extraction. \textsc{Oracle Prompting} engages the type and definition of the extracted target in the task instruction. For example, we query models with the oracle definition \textit{``Verb-particle construction (VPC) is sometimes called phrasal or phrasal-prepositional verb. The meaning of the VPC is fully or partly non-compositional.''} in the extraction procedure targeting an item of VPC. Table \ref{tab:vmwe-oracle-exp} reports the overall accuracy of semantic-enhanced prompting results in \textsc{VMwE} extraction. It is observed that all the models gain from the oracle setting. For instance, Claude-3 increases $+13.5\%$ from $47.1\%$ to $60.6\%$ in accuracy, which indicates that LLMs can be further improved by a more specific target description in the prompt.

\section{Discussion and Takeaways}
We discuss and dissect the research questions predefined in the Section \S\ref{sec:introduction} and the following two.

\paragraph{RQ1: Easy peasy for literal interpretation but still tough nut for generative extraction.}
Comparing the interpretation and extraction of the same type of phrase, we find that all models perform more robustly in the interpretation tasks than in the extraction tasks. Meanwhile, the model performance of generative extraction is still not beyond thorough from that of the human level, especially in the extraction of verbal construction (\textsc{VMwE}), even if prompted with semantic annotation as an oracle that the model was required to follow.  In our tested models, Gemini-Pro achieves the highest accuracy of $77.8\%$ in idiom extraction, while GPT-4 achieves the highest scores of $77.8\%$ and $85.4\%$ in noun compound and collocation, respectively. However, the interpretation tasks achieve high lexical overlap ratio of R-L, and the interpreted meaning exhibits higher fluency than human annotators in our analysis based on textual perplexity (PPL) (See the details in Table \ref{tab:perf-interpretation}). It implies that LLMs with trivial ICL can excel well in the meaning interpretation on \ourbenchmark.

\paragraph{RQ2: Does the winner take it all?}
Regarding LLMs versus SLMs, LLMs have outperformed the trivial supervised models in most tasks. However, as presented in Table \ref{tab:major-exp-results}, SLMs like \texttt{BERT-large} still hold the first tier compared to prompt-based LLMs in IED and LCC. Moreover, as shown in Figure \ref{fig:lcc-eight-classes-cm}, Claude-3 does perform poorly not only in Ver(``genuine'') and Bon(``positive'') but also in their antonyms AntiVer(``non-genuine'') and AntiBon(``negative''). This implies that the semantic discrimination ability of Claude-3 is even worse than BERT's, as reported in \cite{espinosa-anke-etal-2019-collocation}. We conduct the oracle prompting experiment for the more sensitive phrase type \textsc{VMwE}. The results (cf. Table \ref{tab:vmwe-oracle-exp}) reveal that the LLMs can merely be compared with SLMs when we exactly query each of the LLMs with a phrase type and its definition in the extraction procedure. On the other hand, as the number of semantic categories expands, the accuracy of large models diminishes faster than the supervised baseline, following a decay curve with a negative slope. Ultimately, this brings LLMs closer to resembling stochastic parrots \cite{bender2021dangers}. Conversely, without any surprise, fine-tuned small models exhibit greater robustness in accuracy in the same task.

\paragraph{RQ3: Game changers in \textsc{MwE} Processing?}
With the advance of large-scale language modeling, few-shot learning has experienced significant development. Almost all downstream tasks can be achieved within this trend by predicting the next token and yielding outstanding results. Consequently, similar to the situations encountered in other downstream tasks of NLP, we posit that current LLMs have also brought about a paradigm shift in \textsc{MwE} processing. Within this context, the exploration of some research questions such as ``Specialized model vs. LLM, which do we need in \textsc{MwE} processing?'', ``How can NLP systems better handle discontinuous semantic phrases?'' or ``Can large language models serve as the general phrase processing system in some way?'' warrants rethinking and further investigation in the future.

\begin{table}[t]
\renewcommand{\arraystretch}{1.12}
\tiny
\centering
\begin{tabular}{p{1.105cm}cccccccc}
\toprule
\multirow{2.5}{*}{\textbf{System}} & \multicolumn{2}{c}{\textbf{IEI}} & \multicolumn{2}{c}{\textbf{LCI}} & \multicolumn{2}{c}{\textbf{NCI}} \\ \cmidrule{2-7} & \textsc{R-L} $\uparrow$ & \textsc{PPL} $\downarrow$ & \textsc{R-L} $\uparrow$ & \textsc{PPL} $\downarrow$ & \textsc{R-L} $\uparrow$ &  \textsc{PPL} $\downarrow$ \\
\midrule
Human & 24.2 & 34.8 & 26.5 & \uline{\textbf{41.0}} & 39.0 & 65.9 \\
\midrule
\multirow{1}{*}{Gemini-1.0-Pro} & 18.8 & 42.4 & 33.4 & \textbf{\color{ggreen}{62.9}} & 45.2 & 59.1 \\  
${{\hookrightarrow}}\ +$ \textit{3-shot} & 28.2 & 28.7 & 51.4 & 70.8 & 76.0 & 63.1 \\
${{\hookrightarrow}}\ +$ \textit{5-shot} & 27.8 & 28.7 & 50.1 & 66.0 & \textbf{\color{ggreen}{90.0}} & 42.4 \\ \midrule
\multirow{1}{*}{GPT-3.5-Turbo} & \textbf{\color{rred}{14.2}} & 41.3 & 32.5 & 71.5 & 36.3 & 50.3 \\  
${{\hookrightarrow}}\ +$ \textit{3-shot} & 27.4 & 28.0 & 50.5 & 74.9 & 78.0 & 37.1 \\
${{\hookrightarrow}}\ +$ \textit{5-shot} & 27.8 & \textbf{\color{rred}{45.8}} & 50.5 & 74.2 & 81.8 & 39.6 \\ \midrule
 \multirow{1}{*}{GPT-4-Turbo} & 16.3 & 45.2 & 33.1 & \textbf{\color{rred}{97.4}} & \textbf{\color{rred}{28.7}} & \textbf{\color{rred}{72.9}} \\  
${{\hookrightarrow}}\ +$ \textit{3-shot} & 26.8 & 30.8 & 51.0 & 81.0 & 68.5 & \textbf{\color{ggreen}{36.4}} \\
${{\hookrightarrow}}\ +$ \textit{5-shot} & 28.0 & 30.2 & 49.3 & 79.8 & 73.2 & 36.8 \\
 \midrule
 \multirow{1}{*}{Claude-3-Opus} & 21.5 & 34.0 & \textbf{\color{rred}{32.2}} & 75.0 & 50.6 & 41.3 \\  
${{\hookrightarrow}}\ +$ \textit{3-shot} & 27.8 & 26.3 & \textbf{\color{ggreen}{52.6}} & 78.1 & 56.7 & 37.9 \\
${{\hookrightarrow}}\ +$ \textit{5-shot} & \textbf{\color{ggreen}{28.3}} & \textbf{\color{ggreen}{26.2}} & 52.2 & 72.3 & 78.5 & 40.8  \\
\bottomrule
\end{tabular}
\caption{The numerical results, including \textsc{Rouge-L} (R-L), and \textsc{Perplexity} (PPL) of interpretation tasks across the four best-performing models, are shown as an additional report in three interpretation tasks.}
\label{tab:perf-interpretation}
\vspace{-2.1em}
\end{table}

\section{Conclusions}
\label{sec:conclusions}
In this work, we introduced \ourbenchmark, the first benchmark tailored for semantic phrase processing with diverse LMs. We conducted a comprehensive auto-metric evaluation for fifteen models and a human-level comparison across ten tasks. Our findings suggest that notwithstanding assertions of proficiency in surpassing the human performance in most general tasks, there is still much room for improvement of LMs when confronted with the tasks of \ourbenchmark. This discovery underscores the need for more advanced methods to enhance models' capacity to process semantic phrase inputs. Meanwhile, our empirical analysis provides insights into the behavior of LMs in these tasks, thereby guiding future research.

\section*{Limitations}

To our knowledge, our work is the first to systematically explore what extent language models can perform well on a series of semantic phrase processing tasks. However, it has several limitations which are significant to keep in mind and which would be worthwhile to address in future work.

\paragraph{Task Diversity} First, our benchmark involves the representative three kinds of tasks, emit some tasks that also show potential impact on discovering phrasal semantics. Hence, we suggest to construct next generation of phrase comprehension and processing benchmark, to comprise more meaningful tasks on \textsc{MwE}, for instance, semantic retrieval tasks regarding the constituents or entirety of phrases \cite{espinosa-anke-etal-2021-evaluating,pham-etal-2023-pic}.

\paragraph{Phrase Category Coverage} Second, the dataset used in this study includes four kinds of  common phrase phenomena, hence it ignores other semantic phrases which may be thought as a long-tailed distribution. The types of phrase coverage should be wider in the future, in order to enable assessments for more types of semantic phrases. We hope to shed light on this limitation, and spur the \textsc{MwE} community to extend the coverage of \textsc{MwE} beyond the current focus on noun compounds, idioms, and verbal constructions \cite{miletic2024semantics}. The potential extensions include complex function word, multiword named entity and multiword term \cite{constant-etal-2017-survey}.

\paragraph{Multilingual Desiderata} Third, due to the idiosyncrasy, semantic phrases tend to be language-specific. For example, in English or Norwegian, we \textit{take} \text{[\textit{a}]} \textit{nap}, while in Spanish, we \textit{throw} it, and in French, Catalan, German, and Italian, we \textit{make} it \cite{espinosa-anke-etal-2019-collocation}. Our dataset is currently limited to English. Thus, we call for more researchers to participate in dataset design and construction in different languages, towards multilingual resources for a diverse array of semantic phrases.

\section*{Acknowledgements}
We thank Igor Mel'\v{c}uk for his helpful advise on the collocation benchmarking. This work was supported in part by the National Natural Science Foundation of China under Grant 62372039 and in part by the Fundamental Research Funds for the Central Universities under Grant 06500103.

\bibliography{tacl2021v1}
\bibliographystyle{acl_natbib}

\appendix
\section*{Appendix}
\section{Semantic Gloss for Lexical Functions}
\label{sec:appendix-lf_semantic_gloss}
In recent years, there has been an increasing interest in assigning lexical functions as labels to annotated \textsc{MwE} in the sense of the meaning-text theory \cite{mel2023general}. The lexical function is a multi-valued function, which $\mathit{f}$ associates a lexical unit $\mathit{L}$ with a set $\mathit{f(L)}$ of lexical expressions.

As seen in Table \ref{tab:lexfunc-with-semantic-gloss}, we constructed a collection of the representative lexical functions with their semantic glosses from the existing work. We compiled the prompts with the task descriptions.

\begin{table*}
\centering
\scriptsize
\setlength\extrarowheight{1pt}
\begin{tabularx}{\linewidth}{CCC}
\toprule
\textbf{Lexical Function} & \textbf{Semantic Gloss} & \textbf{Complete Description} \\
\midrule
Magn \cite{mel1998collocations} & Intense, strong degree, an intensifier of semantic relation for base lexeme. & Intensify the base lexeme to a high level, strengthening its semantic relation with the associated concept via the collocate lexeme. \\ \hdashline
AntiMagn \cite{mel1998collocations} & Slight and weak degree, a de-intensifier & Weaken meaning intensity, diminishing the semantic relationship between the base lexeme and its associated concept. \\ \hdashline
Ver \cite{gelbukh2012semantic} & Lat. verus, real, genuine & ``As it should be'', ``Meet the intended requirements of \textit{K}''. \\ \hdashline
AntiVer \cite{mel1998collocations} & Non-genuine & Characterize something as non-genuine, not authentic, not in its intended or proper state, and not meeting the required standards or expectations. \\ \hdashline
Bon \cite{espinosa-anke-etal-2021-evaluating} & Positive & Something is good or in a positive situation. \\ \hdashline
AntiBon \cite{espinosa-anke-etal-2021-evaluating} & Negative & Something is bad or in a negative situation. \\ \hdashline
\midrule
IncepPredPlus \cite{fontenelle1997turning} & Start to increase. & Denote initiating a process or action that leads to an increase or enhancement of something. \\ \hdashline
FinFunc0 \cite{kolesnikova2020automatic} & End.existence & The value means ``the \textit{K} of FinFunc0 ceases to be experienced''. \\ \hdashline
Fact0 \cite{mel1998collocations} & Lat. factum, fact. To fulfil the requirement of \textit{K}, and the argument of this function fulfills its own requirement. & Fulfill the base requirement, do something with the base, and do what you are supposed to do with the base. \\
\midrule
CausFunc0 \cite{gelbukh2012semantic} & The agent does something so that the event denoted by the noun occurs & Do something so that \textit{K} begins occurring. \\ \hdashline
Caus1Func0 \cite{espinosa-anke-etal-2021-evaluating} & Cause the existence. 1st argument. & Bring about something's presence or creation, with the first argument indicating the responsible agent or entity. \\ \hdashline
CausFact0 \cite{rodriguez2003domain} & To cause something to function according to its destination. & Denote causing something to function according to its intended purpose or destination. \\ \hdashline
CausPredMinus \cite{fontenelle1997turning} & Cause to decrease. & Describe the act of causing a decrease or reduction in something. \\ \hdashline
CausFunc1 \cite{gelbukh2012semantic} & The non-agentive participant does something such that the event denoted by the noun occurs. & A person/object, different from the agent of \textit{K}, does something so that \textit{K} occurs and has an effect on the agent of \textit{K}. \\ \hdashline
LiquFunc0 \cite{espinosa-anke-etal-2021-evaluating} & Cause termination of the existence & Cause termination of the existence. \\
\midrule
Son \cite{kolesnikova2020automatic} & Lat. \textit{sonare}: sound. & The \textit{K} is usually a noun, and the value means ``emit a characteristic sound''. \\
\midrule
Oper1 \cite{kolesnikova2020automatic} & Lat. \textit{operari}: perform, do, act something. The subject is as the 1st argument. & Represent a light verb linking the event's first participant (subject) with the event's name (direct object). \\ \hdashline
Oper2 \cite{espinosa-anke-etal-2021-evaluating} & Lat. \textit{operari}: perform, do, act something. The subject is as the 2nd argument. & Represent a light verb linking the event's first participant (subject) with the event's name (indirect object). \\ \hdashline
IncepOper1 \cite{gelbukh2012semantic} & Incep is from Lat. \textit{incipere}: begin. Begin to do, perform, experience, carry out \textit{K}. & Signify the start of an action or event, linking the event's subject with its name using a light verb. \\ \hdashline
FinOper1 \cite{kolesnikova2020automatic} & Fin is from Lat. \textit{finire}: cease. & Terminate doing something. \\ \hdashline
Real1 \cite{rodriguez2003domain} & Fulfill a requirement imposed by the noun or performing an action typical for the noun. & To fulfill the requirement of \textit{K}, to act according to \textit{K}. \\
\midrule
Real2 \cite{kolesnikova2020automatic} & Acting as expected. Something be realized as expected  & \textit{K} that is normally expected of the second participant \\ \hdashline
AntiReal2 \cite{kolesnikova2020automatic} & Not acting as expected. Something not be realized as expected. & The \textit{V} is the negation of an internal element of the argument of this function. \\
\bottomrule
\end{tabularx}
\caption{All lexical functions with their semantic gloss in this paper. The column ``semantic gloss'' provides the definition for each LF, and we use a sentence to describe the complete meaning of LF in column ``Complete Description''. \textit{K} denotes the keyword/base word of a LF, and \textit{V} denotes the value/collocate word of a LF.}
\label{tab:lexfunc-with-semantic-gloss}
\end{table*}

\section{Additional Details of Datasets}\label{sec:appendix-additional-experiment-details}

\subsection{Idiomacity Detection}\label{sec:appendix-idiom-detection}
In the initial dataset\footnote{\url{https://github.com/H-TayyarMadabushi/AStitchInLanguageModels}} proposed by \cite{tayyar-madabushi-etal-2021-astitchinlanguagemodels-dataset}, there exists three or four possible meanings (i.e., interpretations) for each instance. For instances with only three interpretations, we add the option ``None of the above'' to keep consistency to the four-choices form. We deduplicate according to the unique ``(idiom, choice)'' pair for all instances. As a result, we collate 273 examples (cf. Table \ref{tab:task-list}). Figure \ref{fig:data-example-idiomacity-detection} shows an example of data.
\begin{figure}[ht]
    \centering
    \fbox{\parbox{0.92\linewidth}{
        [Context] There is also a covered pavilion. It is located next to \textit{\dashuline{Silver Lining}} Tire Recycling. The hours are 6:00 am to 10:00 pm, year round.
        \\ \\
        \text{[Choices]}\\
        (A) grey lining\hfill\redcross \\(B) unexpected advantage\hfill\redcross \\ \textbf{(C) Proper Noun}\hfill\greencheck \\ (D) Meta Usage\hfill\redcross
    }}
    \caption{A data example of idiomacity detection (IED).}
    \label{fig:data-example-idiomacity-detection}
\end{figure}

\subsection{Idiom Extraction}\label{sec:appendix-idiom-extraction}
The original dataset\footnote{\url{https://github.com/Babelscape/ID10M}} consists of instances with or without idiom $\mathcal{IE}$. Since the inference-only experiments comprise most of our work, we filter out all the examples without the $\mathcal{IE}$ existing to increase the coverage diversity of idioms; then, we deduplicate according to the unique item of the occurred $\mathcal{IE}$. The final prepared test set consists of 447 examples with a unique item of $\mathcal{IE}$ existing in each. Figure \ref{fig:data-example-idiom-extraction} shows an example of data.
\begin{figure}[ht]
    \centering
    \fbox{\parbox{0.9\linewidth}{
        [Context] In the screenplay by Lorenzo Semple Jr. , and David Rayfiel , Turner very early on stumbles upon the existence of a kind of super - C.I.A. within the C.I.A. , after which his life is \textit{\dashuline{not worth a plug nickel}} .
        \\ \\
        \text{[Idiom]} \textbf{``not worth a plug nickel''}\hfill\faicon{gavel}
    }}
    \caption{A data example of idiom extraction (IEE).}
    \label{fig:data-example-idiom-extraction}
\end{figure}

\subsection{Idiom Interpretation}\label{sec:appendix-idiom-interpretation}
We collected 916 instances in total from the PIE \cite{zhou-etal-2021-pie}\footnote{\url{https://github.com/zhjjn/MWE_PIE}} and \cite{chakrabarty2022s}\footnote{\url{https://github.com/tuhinjubcse/FigurativeNarrativeBenchmark}}, after deduplication by occurred items of idiom $\mathcal{IE}$. Figure \ref{fig:data-example-idiom-interpretation} shows an example of data.
\begin{figure}[ht]
    \centering
    \fbox{\parbox{0.9\linewidth}{
        \text{[Context]} The remission at this stage of having cancer was truly the \textit{\dashuline{turning point}} of her life .\\\\
        \text{[Idiom]} ``turning point'' \\\\
        \text{[Interpretation]} \textbf{``the time of significant change (mostly positive) in situation''}\hfill\faicon{quote-left}
    }}
    \caption{A data example of idiom interpretation (IEI).}
    \label{fig:data-example-idiom-interpretation}
\end{figure}

\subsection{Lexical Collocation Categorization}\label{subsec:lcc-exp}
We collect the collocation data with the annotated labels from the expanded \textsc{LexFunc}\footnote{\url{https://github.com/luisespinosaanke/lexicalcollocations}} \cite{espinosa-anke-etal-2021-evaluating}. We inherited the training and validation sets of the initial data and sampled 50 examples per semantic category from the test set in classification concerning the computation efficiency. Figure \ref{fig:data-example-collocation-categorization} shows an example of data.
\begin{figure}[ht]
    \centering
    \fbox{\parbox{0.9\linewidth}{
        [Context] In genoa, the \textit{\dashuline{violent storm}} knocked down power lines, blacking out the homes of 5,000 residents.
        \\ \\
        \text{[Category]} \textbf{Magn} (strong semantic).\hfill\faicon{link}
    }}
    \caption{A data example is the lexical collocation categorization (LCC) by semantic relations. Note that the taxonomy included in the prompt is omitted here.}
    \label{fig:data-example-collocation-categorization}
\end{figure}

\subsection{Lexical Collocation Extraction}
The initial dataset is collected from \cite{fisas-etal-2020-collfren}\footnote{\url{https://github.com/TalnUPF/CollFrEn}}. We select the English part of the data and perform deduplication to filter out overlap collocations. We downsample 50 instances randomly for each semantic category to form our test set and reuse the training and validation sets of the original data. Figure \ref{fig:data-example-collocation-extraction} shows an example of data. We conduct $\mathcal{LC}$ extraction but not identification task, and not query models to distinguish the base and the collocate to simplify the task in this work.
\begin{figure}[ht]
    \centering
    \fbox{\parbox{0.9\linewidth}{
        [Context] He still gets up the moment the \textit{\dashuline{alarm clock rings}} .
        \\ \\
        \text{[Semantic relation]} Strong or intense degree in the lexical semantic relation. \\\\
        \text{[Collocation]}\\ \textbf{``alarm clock rings''}\hfill\faicon{map-pin}
    }}
    \caption{A data example of collocation extraction (LCE).}
    \label{fig:data-example-collocation-extraction}
\end{figure}

\subsection{Lexical Collocation Interpretation}\label{sec:appendix-collocation-interpretation}
The data\footnote{\label{footnote4}\url{https://github.com/luisespinosaanke/lexicalcollocations}} we used is proposed in \cite{espinosa-anke-etal-2021-evaluating}. We perform random sampling from the original data and get the 400 examples (50 per class) as our test set. We manually annotated and revised the test examples, and finally get the Cohen's kappa coefficient $\kappa = 0.718$, to confirm the quality. An example of data is shown in Figure \ref{fig:data-example-collocation-interpretation}.

\begin{figure}[ht]
    \centering
    \fbox{\parbox{0.91\linewidth}{
        [Context] Through robert bennett, his lawyer, the president continued friday to call mrs. jones' \textit{\dashuline{baseless accusation}}.
        \\ \\
        \text{[Collocation]} ``baseless accusation'' \\ \\
        \text{[Interpretation]} \textbf{``Groundless claim made without substantiation''}\hfill\faSearch
    }}
    \caption{A data example of collocation interpretation (LCI).}
    \label{fig:data-example-collocation-interpretation}
\end{figure}

\subsection{Noun Compound Compositionality}
\begin{figure}[ht]
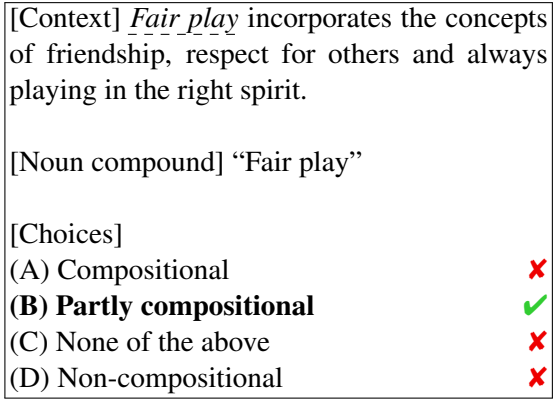

    \centering
    \fbox{\parbox{0.92\linewidth}{
        [Context] \textit{\dashuline{Fair play}} incorporates the concepts of friendship, respect for others and always playing in the right spirit.
        \\ \\
        \text{[Noun compound]} ``Fair play''
        \\ \\
        \text{[Choices]}\\
        (A) Compositional\hfill\redcross \\\textbf{(B) Partly compositional}\hfill\greencheck \\ (C) None of the above\hfill\redcross \\ (D) Non-compositional\hfill\redcross
    }}
    \caption{A data example of noun compound compositionality (NCC).}
    \label{fig:data-example-noun-compound-compositionality}
\end{figure}

The annotated noun compound data is collected from the NCTTI\footnote{\url{https://github.com/marcospln/nctti}} \cite{garcia-etal-2021-assessing}. After data processing, we filtered out the compound without reference context, collated 237 examples, and split them into training, validation, and test sets. Figure \ref{fig:data-example-noun-compound-compositionality} shows an example of data.

\subsection{Noun Compound Extraction}
\begin{figure}[ht]
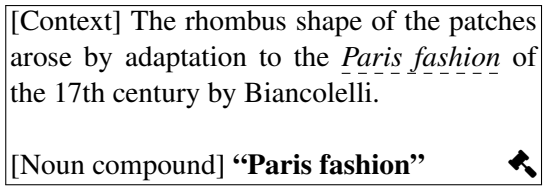

    \centering
    \fbox{\parbox{0.9\linewidth}{
        [Context] The rhombus shape of the patches arose by adaptation to the \textit{\dashuline{Paris fashion}} of the 17th century by Biancolelli.
        \\ \\
        \text{[Noun compound]} \textbf{``Paris fashion''}\hfill\faicon{gavel}
    }}
    \caption{A data example of noun compound extraction (NCE).}
    \label{fig:data-example-noun-compound-extraction}
\end{figure}

As our beginning, we sampled the test set from the \textsc{ProNCI}\footnote{\url{https://github.com/dair-iitd/pronci}} \cite{kolluru-etal-2022-covid}. We used the training and validation sets to leverage the compositional part of noun compounds in the original dataset. We randomly sampled from the test set to form the new test set with 720 examples. We demonstrate a data example in Figure \ref{fig:data-example-noun-compound-extraction}.

\subsection{Noun Compound Interpretation}
\begin{figure}[ht]
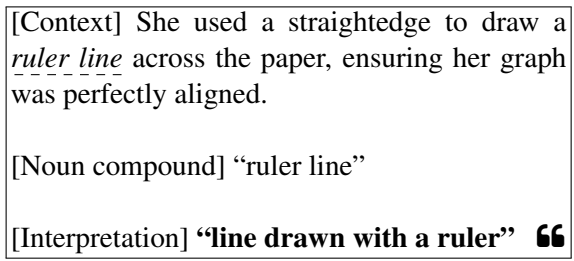

    \centering
    \fbox{\parbox{0.95\linewidth}{
        \text{[Context]} She used a straightedge to draw a \textit{\dashuline{ruler line}} across the paper, ensuring her graph was perfectly aligned. \\\\
        \text{[Noun compound]} ``ruler line'' \\\\
        \text{[Interpretation]} \textbf{``line drawn with a ruler''}~~~\faicon{quote-left}
    }}
    \caption{A data example of noun compound interpretation (NCI).}
    \label{fig:data-example-noun-compound-interpretation}
\end{figure}

We leverage the initial training, validation, and test data splits from \cite{coil-shwartz-2023-chocolate}\footnote{\url{https://github.com/jordancoil/noun-compound-interpretation}}. To provide a context for each noun compound, we use ChatGPT to generate a reference sentence. To verify the quality of synthetic data, we performed a manual inspection, which resulted in  $acc>98\%$. A data example is shown in the figure \ref{fig:data-example-noun-compound-interpretation}.

\subsection{\textsc{VMwE} Extraction}
\begin{figure}[ht]
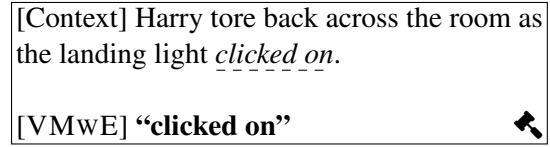

    \centering
    \fbox{\parbox{0.9\linewidth}{
        [Context] Harry tore back across the room as the landing light \textit{\dashuline{clicked on}}.
        \\ \\
        \text{[\textsc{VMwE}]} \textbf{``clicked on''}\hfill\faicon{gavel}
    }}
    \caption{A data example of \textsc{VMwE} Extraction.}
    \label{fig:data-example-vmwe-extraction}
\end{figure}

We used the English corpus of PARSEME v1.3\footnote{\url{https://gitlab.com/parseme/parseme_corpus_en}} \cite{savary-etal-2023-parseme}, the existing largest annotated corpora of \textsc{VMwE}. The initial data is used to conduct extraction instead of identification tasks. Figure \ref{fig:data-example-vmwe-extraction} shows an example of the data.

\section{Example Prompt}\label{sec:example-prompt}

We manually create a unified prompt template for all tasks that can be adapted to each task with specific filling arguments. The prompt format is shown in the Figure \ref{unified-prompt-template}. The detailed prompt for each task can be accessed in our code base\footnote{\url{https://github.com/jacklanda/LexBench/exp/prompts}}. \\

\begin{center}
	\scalebox{0.8}{
		\begin{tcolorbox}[enhanced,attach boxed title to top center={yshift=-3mm,yshifttext=-1mm},
		  colback=white,colframe=black!75!black,colbacktitle=black!80!black,
		  title=Unified Prompt Template,fonttitle=\bfseries\small, boxed title style={size=small,colframe=black!50!black}, fontupper=\footnotesize]
		  \em Assume that you are a linguistic who research in the \text{\{\{semantic phrase\}\}}. \\ \\
		  You will be given a sentence that contains only an item of \text{\{\{semantic phrase\}\}}. \\ \\
		  Your task is to ... \\ \\
		  Please make sure you read and understand these instructions carefully. \\
		  \begin{DottedTextBox}
		    Few-shot Examples: \\
		    \begin{DottedTextBox}Phrase: \textup{\{\{}an example of the phrase\textup{\}\}} \end{DottedTextBox}
		    Context: \textup{\{\{}a context of the example\textup{\}\}} \\
		    Output: \textup{\{\{}an output of the example\textup{\}\}} \\
		    \textbf{...}
		  \end{DottedTextBox}
		  \begin{DottedTextBox}Phrase: \textup{\{\{}phrase\textup{\}\}}\end{DottedTextBox}
		  Context: \textup{\{\{}context\textup{\}\}} \\ \\
		  Output: \\
		\end{tcolorbox}
	}
\end{center}
\begin{center}
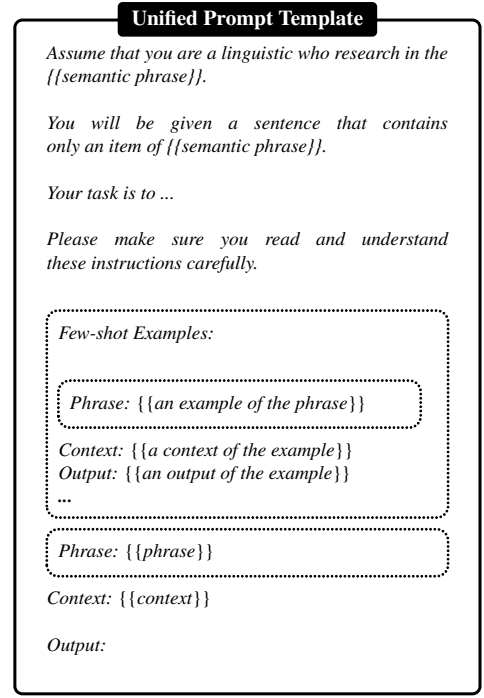

    \begin{minipage}{\linewidth}
        \centering
        \captionof{figure}{Unified prompt template used in the work.}\label{unified-prompt-template}
    \end{minipage}
\end{center}

\section{Additional Experiment Details}\label{sec:additional-exp-details}
\begin{table}[t]
    \centering
    \tiny
    \begin{tabular}{l@{\hspace{0.8em}}c@{\hspace{0.5em}}c@{\hspace{0.8em}}c@{\hspace{0.9em}}c@{\hspace{1em}}c}
        \toprule
            \textbf{Model} & \textbf{\# Params} & \textbf{Arch.} & \textbf{Creator} & \textbf{Public} & \textbf{Post Training} \\
        \midrule
            \hlcelll BERT base$^\dagger$ & 110M & Enc. & Google & \greencheck & SFT \\
            \hlcelll BERT large$^\dagger$ & 340M & Enc. & Google & \greencheck & SFT \\
            \hlcelll T5 base$^\dagger$ & 220M & Enc.+Dec. & Google &  \greencheck & SFT \\
            \hlcelll T5 large$^\dagger$ & 770M & Enc.+Dec. & Google & \greencheck & SFT \\
        \midrule
            \hlcellll Llama 2 7B chat & 7B & Dec. & Meta & \greencheck & SIFT \\
            \hlcellll Llama 2 13B chat & 13B & Dec. & Meta & \greencheck & SIFT \\
            \hlcellll Llama 2 70B chat & 70B & Dec. & Meta & \greencheck & SIFT \\
            \hlcellll DeepSeek 7B chat & 7B & Dec. & DeepSeek & \greencheck & SIFT \\
            \hlcellll Mistral 7B inst & 7B & Dec. & Mistral AI & \greencheck & SIFT \\
            \hlcellll Mixtral 8x7B inst & 46.7B & Dec.(MoE) & Mistral AI &  \greencheck & SIFT \\
        \midrule
            \hlcell GPT-3.5 turbo$^\ddagger$ & * & * & OpenAI & \redcross & SIFT + RLHF \\
            \hlcell GPT-4 turbo$^\ddagger$ & * & * & OpenAI & \redcross & SIFT + RLHF \\
            \hlcell Claude instant 1$^\ddagger$ & * & * & Anthropic & \redcross & SIFT + RLHF \\
            \hlcell Claude 3 opus$^\ddagger$ & * & * & Anthropic & \redcross & SIFT + RLHF \\
            \hlcell Gemini 1.0 pro$^\ddagger$ & * & * & Google  & \redcross & SIFT + RLHF \\
        \bottomrule
    \end{tabular}
    \caption{
        A list of LMs tested in this paper:
        ``Public'' indicates whether the model weights are open. In detail, \colorbox{sred}{Light Pink} text delineates the supervised fine-tuned models. \colorbox{sgreen}{Light Green} and \colorbox{sblue}{Light Blue} parts present open-source models and proprietary models, respectively.
        ``Post Training'' indicates whether the model is trained further in some ways after pre-training.
        $^\dagger$We perform trivial full-set fine-tuning for the models.
        $^\ddagger$We use the official API for the model inference.
    }
    \label{tab:models}
    \vspace{-1.4em}
\end{table}
For the sequence classification tasks such as LCC, we employ \texttt{bert-base/large-uncased} as our tuning initiation. Similarly, we construct primary baselines for extraction tasks that leverage the B-I-O scheme to conduct sequence labeling. The training is run with an NVIDIA A100-40GB on Google Colab \cite{bisong2019google}. For interpretation tasks, we use \texttt{t5-base/large} model to conduct vanilla fine-tuning. Additionally, We train all models for a specific number of epochs shown in Table \ref{tab:hyperparams-bert-t5} and perform early stopping over the validation set. Model checkpoints used in our experiment are implemented by PyTorch \cite{paszke2019pytorch}, and Hugging Face Transformers \cite{wolf-etal-2020-transformers}. The input format of the prompt and the few-shot demonstration settings we used during the experiment are shown in Figure \ref{unified-prompt-template}. Since each model has different generation styles, we conduct a pre-run before each test. Then, we develop ad hoc heuristics based on the response generated by models to parse predictions accurately. The perplexity computing in the interpretation tasks is to feed the phrase and its interpretation into the template \textit{``The meaning of phrase \{\{phrase\}\} in context is \{\{interpretation\}\}''}, and then we compute the token-level perplexity by GPT-2-XL \cite{radford2019language}.

\section{Annotation Guideline}\label{sec:annotation-guideline}
We established the following criteria for compiling the dataset of collocation interpretation (\S\ref{subsec:Collocation-Interpretation}).
\begin{enumerate}
  \item \textbf{Objective:} Interpret each lexical collocation in five distinct narratives for comprehensive understanding according to the given context.
  \item \textbf{Dataset Overview:} Contains context and collocations paired with  base and collocate.

  \item \textbf{Annotation Format:} Include collocation, five narratives (N1-N5), and rationale.
  
  \item \textbf{Consistency and Accuracy:} Maintain consistent and accurate interpretations across the five narratives in the same semantic meaning.
\end{enumerate}

\begin{table*}
\centering
\small
\setlength\tabcolsep{4pt}
\setlength{\extrarowheight}{3pt}
\begin{tabular}{lccccccccccc}
\toprule
\multirowcell{2.4}{\textbf{\textsc{VMwE}}} & \phantom & \multicolumn{3}{c}{\textbf{BERT-base}} & \phantom{c} & \multicolumn{3}{c}{\textbf{BERT-large}} & \phantom & \multirowcell{2.4}{\textbf{\# Support}} & \phantom{c} \\
\cmidrule{3-5}\cmidrule{7-9}
&& \textbf{P} & \textbf{R} & \textbf{F1} && \textbf{P} & \textbf{R} & \textbf{F1} \\
\midrule
IAV && 60.74$_{\text{5.59}}$ & 37.96$_{\text{4.25}}$ & 46.53$_{\text{3.27}}$ && 46.51$_{\text{3.57}}$ & 38.89$_{\text{5.56}}$ & 42.33$_{\text{4.78}}$ && 36 \\
LVC.cause && 46.41$_{\text{12.21}}$ & 18.39$_{\text{3.98}}$ & 26.23$_{\text{5.58}}$ && 26.43$_{\text{12.42}}$ & 20.69$_{\text{9.13}}$ & 23.2$_{\text{10.53}}$ && 29 \\
LVC.full && 52.12$_{\text{4.35}}$ & 61.05$_{\text{2.01}}$ & 56.12$_{\text{2.21}}$ && 55.23$_{\text{2.48}}$ & 56.78$_{\text{8.41}}$ & 55.88$_{\text{5.42}}$ && 172 \\
MVC && 95.94$_{\text{4.00}}$ & 80.46$_{\text{1.99}}$ & 87.52$_{\text{2.76}}$ && 100.0$_{\text{0.00}}$ & 80.46$_{\text{1.99}}$ & 89.16$_{\text{1.22}}$ && 29 \\
VID && 52.4$_{\text{5.34}}$ & 36.11$_{\text{0.93}}$ & 42.67$_{\text{1.79}}$ && 63.83$_{\text{5.04}}$ & 36.11$_{\text{1.85}}$ & 46.07$_{\text{2.01}}$ && 108 \\
VPC.full && 64.28$_{\text{3.28}}$ & 78.35$_{\text{1.55}}$ & 70.56$_{\text{1.47}}$ && 64.35$_{\text{0.08}}$ & 79.38$_{\text{0.52}}$ & 71.07$_{\text{0.22}}$ && 194 \\
VPC.semi && 55.93$_{\text{38.65}}$ & 8.89$_{\text{6.94}}$ & 12.9$_{\text{7.30}}$ && 38.78$_{\text{6.41}}$ & 35.55$_{\text{3.85}}$ & 37.08$_{\text{5.01}}$ && 30 \\
\midrule
Micro Avg. && 63.24$_{\text{1.94}}$ & 61.2$_{\text{0.58}}$ & 62.3$_{\text{1.32}}$ && 64.19$_{\text{0.46}}$ & 62.37$_{\text{2.66}}$ & 63.26$_{\text{1.55}}$ && 85.42\\
\bottomrule
\end{tabular}
\caption{We report the full results of \textsc{VMwE} extraction reproduced on MTLB-STRUCT. The performance of all categories are defined in the corpora PARSEME 1.3. The corresponding standard deviation is calculated by the results of three runnings with the selected seeds $\{21, 42, 84\}$.} 
    \label{tab:vmwe-full-results}
\end{table*}

\begin{table*}[!ht]
    \centering
    \small

    \begin{tabular}{c}
      \toprule
      \textbf{Computing Infrastructure}\\1\ $\times$\ A100 40GB GPU (Google Colab) \\ 
      \bottomrule \\
    
    \vspace{3mm}\begin{tabular}{cc}
        \toprule
        \textbf{Hyperparameter} & \textbf{Assignment}  \\
        \midrule
        architecture & BERT-\{base, large\} \\
        \midrule
        tokens per sample & $150$ \\
        \midrule
        batch size & $4,800$ \\
        \midrule
        number of workers & $8$ \\
        \midrule
        learning rate & $3e^{-5}$ \\
        \midrule
        number of epochs & $10$ \\
        \midrule
        save interval (epoch) & $1$ \\
        \midrule
        validation interval (epoch) & $1$ \\
        \midrule
        ratio of warmup steps & $3\%$ \\
        \midrule
        learning rate scheduler & Polynomial decay \\
        \midrule
        learning rate optimizer & Adam \\
        \midrule
        Adam beta weights & $(0.9, 0.99)$ \\
        \midrule
        Adam epsilon & $1e^{-6}$ \\
        \midrule
        weight decay & $0$ \\
        \midrule
        random seed & $21$, $42$, $84$ \\
        \bottomrule
    \end{tabular}
    \vspace{3mm}\begin{tabular}{cc}
        \toprule
        \textbf{Hyperparameter} & \textbf{Assignment}  \\
        \midrule
        architecture & T5-\{base, large\} \\
        \midrule
        tokens per sample & $128$ \\
        \midrule
        batch size & $2,048$ \\
        \midrule
        number of workers & $4$ \\
        \midrule
        learning rate & $5e^{-5}$ \\
        \midrule
        number of epochs & $5$ \\
        \midrule
        save interval (epoch) & $1$ \\
        \midrule
        validation interval (epoch) & $1$ \\
        \midrule
        ratio of warmup steps & $3\%$ \\
        \midrule
        learning rate scheduler & Cosine decay \\
        \midrule
        learning rate optimizer & Adam \\
        \midrule
        Adam beta weights & $(0.9, 0.99)$ \\
        \midrule
        Adam epsilon & $1e^{-6}$ \\
        \midrule
        weight decay & $0$ \\
        \midrule
        random seeds & $21$, $42$, $84$ \\
        \bottomrule
    \end{tabular}
    \vspace{-3mm}
    \\ \bottomrule
    \end{tabular}
    
    \caption{Hyperparameters for finetuning BERT-Taggers and T5 Generators.} 
    \label{tab:hyperparams-bert-t5}
\end{table*}

\begin{table*}
\centering
\footnotesize
\setlength\tabcolsep{4pt}
\setlength{\extrarowheight}{2pt}
\begin{tabular}{lccccc}
\toprule
\textbf{System} & \textbf{Acc@1} & \textbf{Acc@2} & \textbf{Acc@4} & \textbf{Acc@8} & \textbf{Acc@16} \\
\midrule
\underline{\textbf{\textit{Baselines}}} & & & & & \\
Random & 100.00 & 50.00 & 25.00 & 12.50 & 6.25 \\
Majority & 100.00 & 50.00 & 25.00 & 12.50 & 6.25 \\
\midrule
\underline{\textbf{\textit{Small language models}}} & & & & & \\
BERT$_\text{B}$ & 100.00$_\text{0.00}$ & 98.88$_\text{1.92}$ & 89.44$_\text{4.88}$ & 79.86$_\text{6.40}$ & 69.93$_\text{2.48}$ \\
BERT$_\text{L}$ & 100.00$_\text{0.00}$ & \textbf{98.88$_\text{0.96}$} & \textbf{95.83$_\text{1.44}$} & \textbf{83.47$_\text{5.17}$} & \textbf{71.80$_\text{1.02}$} \\
\midrule
\underline{\textbf{\textit{Large language models}}} & & & & & \\
GPT-3.5-turbo-0613 & 100.00$_\text{0.00}$ & 63.89$_\text{13.46}$ & 45.28$_\text{20.75}$ & 28.61$_\text{14.56}$ & 17.69$_\text{4.89}$ \\
~~~$\hookrightarrow +$ \textit{3-shot} & 100.00$_\text{0.00}$ & 66.66$_\text{15.27}$ & 49.16$_\text{14.01}$ & 37.08$_\text{12.21}$ & 24.72$_\text{1.61}$ \\
~~~$\hookrightarrow +$ \textit{5-shot} & 100.00$_\text{0.00}$ & 72.22$_\text{11.82}$ & 51.39$_\text{15.14}$ & 39.16$_\text{11.32}$ & 26.94$_\text{0.78}$ \\
GPT-4-1106-preview & 100.00$_\text{0.00}$ & 82.78$_\text{11.34}$ & 60.55$_\text{15.73}$ & 40.27$_\text{7.88}$ & 34.17$_\text{1.46}$ \\
~~~$\hookrightarrow +$ \textit{3-shot} & 100.00$_\text{0.00}$ & 87.78$_\text{9.76}$ & 70.00$_\text{10.4}$ & 52.5$_\text{12.2}$ & 42.98$_\text{1.61}$ \\
~~~$\hookrightarrow +$ \textit{5-shot} & 100.00$_\text{0.00}$ & 89.34$_\text{7.11}$ & 70.83$_\text{9.27}$ & 54.1$_\text{8.32}$ & 45.07$_\text{1.58}$ \\
Claude-3-opus & 100.00$_\text{0.00}$ & 90.55$_\text{12.28}$ & 72.77$_\text{12.75}$ & 49.58 $_\text{5.90}$ & 39.30 $_\text{0.98}$ \\
~~~$\hookrightarrow +$ \textit{3-shot} & 100.00$_\text{0.00}$ & 90.55$_\text{10.71}$ & 75.83$_\text{6.66}$ & \underline{58.47$_\text{7.56}$} & \underline{54.1$_\text{8.32}$} \\
~~~$\hookrightarrow +$ \textit{5-shot} & 100.00$_\text{0.00}$ & \underline{90.55$_\text{9.18}$} & \underline{76.94$_\text{8.91}$} & 57.5$_\text{7.78}$ & 46.25$_\text{1.78}$ \\
Gemini-1.0-pro & 100.00$_\text{0.00}$ & 84.44$_\text{10.58}$ & 54.16$_\text{7.40}$ & 32.08$_\text{6.88}$ & 24.65$_\text{1.32}$ \\
~~~$\hookrightarrow +$ \textit{3-shot} & 100.00$_\text{0.00}$ & 90.00$_\text{8.81}$ & 66.38$_\text{14.82}$ & 43.33$_\text{9.38}$ & 36.11$_\text{0.52}$ \\
~~~$\hookrightarrow +$ \textit{5-shot} & 100.00$_\text{0.00}$ & 90.00$_\text{11.54}$ & 65.0$_\text{13.91}$ & 44.85$_\text{10.00}$ & 37.43$_\text{1.41}$ \\
\bottomrule
\end{tabular}
\caption{Our best experimental results (avg$_\text{std}$). The mean accuracy scores with their standard deviation are computed by averaging the results of three independent runs with different random seeds. Results of baselines are also provided including random choice as well as the majority of class instances over each sub categorization tasks. The \textbf{Bold} and \underline{underlined} texts denote the best and second-best performance in the specific category, respectively.}
\label{tab:lcc-scaling-exp-detail}
\end{table*}

\end{document}